%% file: main.tex
\documentclass[10pt,journal,compsoc]{IEEEtran}

\ifCLASSOPTIONcompsoc
  \usepackage[nocompress]{cite}
\else
  \usepackage{cite}
\fi

\ifCLASSINFOpdf
\else
\fi

\usepackage[table]{xcolor}
\usepackage{times}
\usepackage{epsfig}
\usepackage{graphicx}
\usepackage{amsmath}
\usepackage{amssymb}
\usepackage{booktabs}
\usepackage{color}
\usepackage{float}
\usepackage{array}
\usepackage{multirow}
\usepackage{makecell}
\usepackage{amssymb}
\usepackage{caption}
\usepackage{soul}
\usepackage{xcolor}
\usepackage{cuted}
\usepackage{capt-of}
\usepackage[OT1]{fontenc}
\usepackage{paralist}
\usepackage{bm}

\usepackage{mathrsfs}
\usepackage{color}
\usepackage{booktabs}
\usepackage{multirow}
\usepackage{threeparttable}
\usepackage{color, colortbl}
\usepackage{changes}
\usepackage{amssymb}
\usepackage{graphicx}
\usepackage{amsmath}
\usepackage{ragged2e}
\usepackage{setspace}

\usepackage[super]{nth}
\usepackage[pagebackref=true,colorlinks,bookmarks=false,citecolor=blue,linkcolor=blue,urlcolor=blue]{hyperref}

\definecolor{LightCyan}{rgb}{0.5,1,1}
\definecolor{mypink3}{cmyk}{0, 0.7808, 0.4429, 0.1412}

\input{mlVecMat}

\usepackage[capitalize]{cleveref}
\crefname{section}{Sec.}{Secs.}
\Crefname{section}{Section}{Sections}
\Crefname{table}{Table}{Tables}
\crefname{table}{Tab.}{Tabs.}

\def\eg{\emph{e.g}.}

\definecolor{lightgray}{gray}{0.9}
\usepackage[switch]{lineno}

\begin{document}
%
\title{High-Quality Entity Segmentation and Grounding}
%
%
%
%

\author{Lu Qi,
        Yi-Wen Chen,
        Tao Zhang,
        Xiangtai Li,
        Xu Yang,
        Bo Du,
        Ming-Hsuan Yang

\IEEEcompsocitemizethanks{\IEEEcompsocthanksitem Lu Qi, Tao Zhang, and Bo Du are with Wuhan University. Lu Qi is also affiliated with Insta360 Research. Yi-Wen Chen and Ming-Hsuan Yang are with the Department of EECS, University of California, Merced. Xiangtai Li and Xu Yang are with Nanyang Technological University, and the Institute of Automation of the Chinese Academy of Sciences. Xiangtai Li and Xu Yang are the corresponding authors.}
}

\IEEEtitleabstractindextext{
\justifying
\begin{abstract}
Class-agnostic segmentation is of great importance for generalizing the localization of semantically-consistent regions from limited training data to unseen domains and categories.
It has benefited numerous applications, such as human-interaction-assisted image editing.
%
Despite their success, existing class-agnostic methods (e.g., SAM2 and entity segmentation) still struggle with boundary-accurate masks and semantic understanding.
To address these issues, we propose ESG, a pipeline for high-quality entity segmentation and grounding supported by a new dataset EntitySeg.
%
%
At first, the proposed dataset naming EntitySeg contains images spanning various image domains and entities, along with plentiful high-resolution images and high-quality mask annotations for training and testing. 
Then, the ESG mainly consists of two modules: CropFormer for high-quality entity segmentation whereas GELLA for accurate noun extraction from sentences and semantic matching between language and visual regions.
%
%
Unlike existing grounding methods that jointly train a segmentation and a large language model, ESG adopts a two-stage decoupled design, preserving high-quality masks and grounding robustness without the trade-offs often introduced by joint training.
CropFormer ensures high-quality entity segmentation results, which can then be encoded into the GELLA model for effective grounding.
Extensive experimental results demonstrate the effectiveness of our proposed pipeline across five tasks, including entity segmentation, panoptic segmentation, open-vocabulary segmentation, referring segmentation, and panoptic localized narratives.
Furthermore, GELLA module of ESG pipeline is highly flexible and capable of processing mask inputs from any segmentation framework, thanks to its lightweight colormap/vision encoder, language/mask decoder, and association module.
The entity segmentation dataset and grounding code will be released at \href{https://github.com/qqlu/Entity}{https://github.com/qqlu/Entity}.
\end{abstract}

}

\maketitle

\IEEEdisplaynontitleabstractindextext

%
\IEEEpeerreviewmaketitle

\IEEEraisesectionheading{\section{Introduction}\label{sec:introduction}}

\IEEEPARstart{C}{lass-agnostic} segmentation~\cite{kirillov2023segment,qi2021open}, is a promising vision task for pixel-level localization without relying on semantic labels, enabling strong generalization to semantic-consistent regions that were even unseen in the training data.
Recent applications usually leveraged interactive class-agnostic segmentation for image manipulation~\cite{wang2022manitrans, wang2023entity} and retouching~\cite{liu2022text}, where users can specify any text queries (e.g., Oculus, platypus) for local editing. 
Although the interactive models~\cite{kirillov2023segment} have achieved great success, their interactive pipeline, low-quality partial mask annotations and unawareness of categories block their development for broader areas, such as anomaly detection in autonomous cars, which require an automatic, accurate, and consistent dense prediction wrapped with semantic understanding.

\begin{figure*}[tp]
  \centering
  \includegraphics[width=0.9\linewidth]{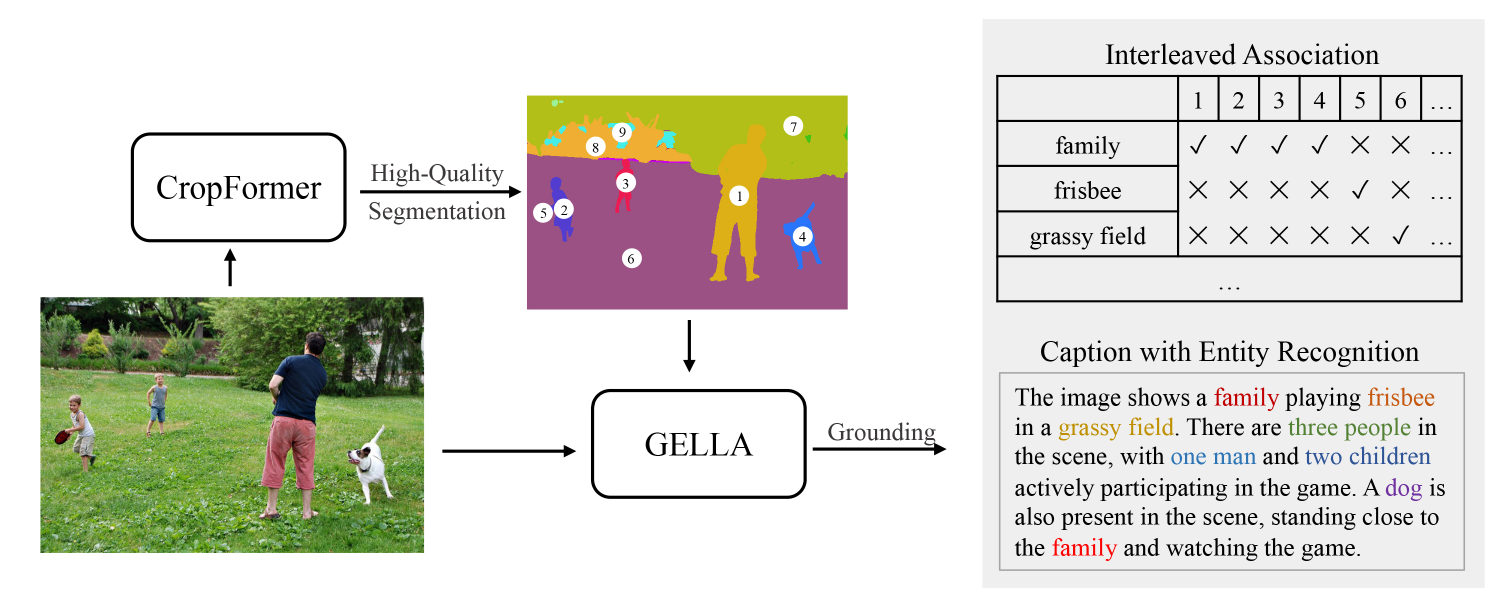}
  \caption{\textbf{Overview of our proposed ESG pipeline for high-quality entity segmentation and grounding.} With the help of a large language model, the grounding includes an image-level caption with entity recognition and interleaved association.}
  \label{fig:gella_overview}
\end{figure*}

The alternative is to associate entity segmentation and grounding ability of large language model.
However, this association still faces two challenges. On one hand, segmentation quality remains limited due to the lack of high-quality segmentation data and effective methods. On the other hand, semantic grounding must be integrated with segmentation in a unified framework, while keeping inference efficiency to the number of entities.
%
%
%
%

%
To address these issues, we propose the ESG pipeline for high-quality \textbf{E}ntity \textbf{S}egmentation and \textbf{G}rounding. As shown in Figure~\ref{fig:gella_overview}, ESG mainly consists of three core components: CropFormer, along with a new high-quality entity segmentation dataset, EntitySeg, for robust entity segmentation; and GELLA for accurate entity grounding via assistance of large language model. Such a decoupled design is inspired by the standard two-stage Faster R-CNN that includes the region proposal network (RPN) and R-CNN and can perform better than the joint training pipeline for multiple tasks.
CropFormer guarantees high-quality entity segmentation results that can be encoded in the GELLA model for grounding.

\begin{figure*}[t!]
  \centering
  \includegraphics[width=1.0\linewidth]{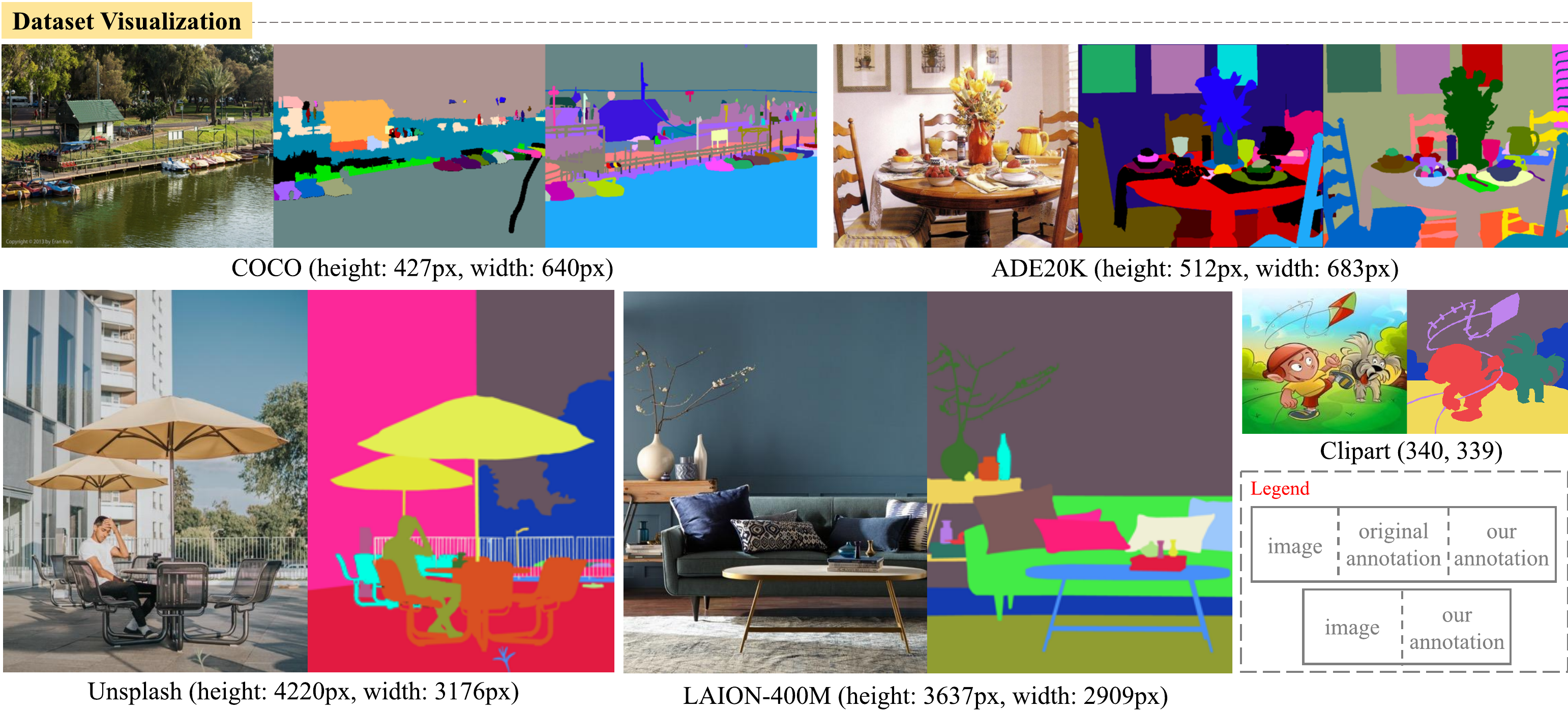}
  \caption{\textbf{Fine-grained entity segmentation mask annotations for low-resolution and high-resolution images collected from existing datasets and the Internet.} 
  For images collected from COCO~\cite{lin2014microsoft} and ADE20K~\cite{zhou2017scene}, we provide visual comparisons between their original and our annotations in the middle and rightmost sub-figures. 
  In existing datasets, the areas in black represent unannotated regions. 
  Please note that the RGB and mask images shown here have been down-sampled, which has degraded their quality compared to the actual dataset. 
  We denoted pixels as px for simplicity.}
  \label{fig:comp_coco}
\end{figure*}

%
%
For the EntitySeg dataset, we prioritize high-quality mask annotations and high-resolution images, essential for the current era of high-resolution media.
%
Compared to the existing segmentation datasets containing low-resolution images
(\eg, COCO~\cite{lin2014microsoft} and SA1B~\cite{kirillov2023segment}) do not have images that are above 1.6K image size and coarsely annotated masks, the EntitySeg dataset makes it easy to label more accurate mask annotations by collecting higher-resolution images whose average image size surpasses 2.8K.
As such, one-third of images are annotated with a global-level caption by GPT-4O and entity-level open-vocabulary tags by annotators.
The Figure~\ref{fig:comp_coco} shows the annotation quality of sampled images in our EntitySeg dataset and the comparison to other popular segmentation datasets such as COCO~\cite{lin2014microsoft} and ADE20K~\cite{zhou2017scene}.
%
%
%
%

%
Then, the region proposal network of ESG, the CropFormer, learns to dynamically generate tokens to associate the full image with its high-resolution crops, fusing them to generate the final class-agnostic mask prediction.
During training, one of the four predefined corner crops is randomly selected and fused with the full image. 
CropFormer is applied to all four corner crops and the full image for complete fusion during inference. 
CropFormer's strength lies in its ability to dynamically generate queries linking the same entities in the full image and its higher-resolution crops. This allows us to benefit from both the global image context and high-quality image/annotation details to obtain improved results. 
%

Based on high-quality entity segmentation predictions, our proposed ESG pipeline adopts the GELLA framework for \textbf{G}eneralizable \textbf{E}ntity grounding with \textbf{L}arge \textbf{L}anguage \textbf{A}ssistance.
Unlike other methods that rely on online mask generation, GELLA in ESG pipeline directly takes the segmentation masks produced by CropFormer as input, making the design more generalizable to other segmentation architectures.
Specifically, we assign a unique random color to each entity on a colormap, allowing encoded features to provide a robust mask prior to a fixed lightweight structure. 
Thus, the colormap encoder significantly reduces the computational workload required for pixel-level prediction and enhances framework flexibility to accommodate various colormaps. 
This means that our framework can be further improved with the development of each pre-trained model.
%
In the GELLA module, we propose a ResoBlend module to merge features from the image and mask extracted from a low-resolution CLIP vision and colormap encoder.
The fused features are then used to accurately reconstruct the original segmentation masks in conjunction with the entity embeddings.
These entity embeddings harmonize more effectively with the language embeddings, as both are derived from the CLIP vision encoder, ensuring a more consistent association between the visual and linguistic elements.
%

We conduct extensive experiments on five tasks, including referring expression segmentation, panoptic narrative grounding, panoptic segmentation, reasoning segmentation, and open-vocabulary segmentation. 
With the curated EntitySeg dataset and proposed modules of CropFormer and GELLA, the ESG pipeline outperforms or is comparable to state-of-the-art methods while retaining the ability to interact with users.
Compared to the strategy of cropping the mask region and pooling features, our design preserves the context features and positional information across the entire feature map.
This enables more accurate association, especially when multiple entities belong to the same category, as proved by the experiment on open-vocabulary segmentation.

The main contributions of proposed ESG pipeline for high-quality entity segmentation and grounding in this work are:
\begin{itemize}
    \item We build a large-scale, high-quality entity segmentation dataset, EntitySeg, with images collected from various domains with different resolutions and high-quality masks.
    The EntitySeg enables the evaluation of the generalization and robustness capabilities of segmentation models in a unified manner through an domain-diverse and high-resolution setting.
    \item
    We benchmark several segmentation models on the EntitySeg dataset. 
    We find that the conventional training setting cannot fit well with our dataset, where over 70\% are very high in resolution with high computational costs. 
    To solve this problem, we propose CropFormer that
    associates the same entities in different crops with the same queries.
    To the best of our knowledge, this is the first ensemble-based prediction using different image crops in query-based segmentation frameworks. 
    Extensive experiments and in-depth analysis show the merits of the CropFormer for high-quality entity segmentation.

    \item Based on the results of the entity segmentation from CropFormer, the ESG pipeline further utilize proposed GELLA module for generalizable entity grounding with a user prompt. 
    Specifically, we leverage an LLM to associate semantic nouns in the caption with class-agnostic entity segmentation masks using two single-modal encoders and two dual-modal decoders. 
    We have a colormap encoder that enables the preservation of fine-grained predictions from high-resolution masks.
    This encoder allows us to extract visual features from low-resolution images using a CLIP vision encoder, significantly reducing computational costs. 
    Extensive experiments of the ESG on panoptic narrative grounding, referring expression segmentation, open/closed panoptic segmentation, and open-vocabulary segmentation demonstrate the effectiveness of the proposed method. 
    Given its flexibility to accept multimodal input, ESG can consistently leverage state-of-the-art single-modal methods.
\end{itemize}

\vspace{1mm}
\noindent \textbf{Difference from our conference paper.} This manuscript is an extension to the high-quality entity segmentation paper~\cite{qi2023high} that was presented in ICCV 2023.
While the CropFormer was first introduced in our ICCV paper, this work focuses on ground CropFormer’s segmentation results using the extended ESG pipeline.
The ESG pipeline achieves its grounding objectives through proposed GELLA module: it generates image captions, extracts semantic nouns from these captions, and associates these nouns with the segmentation masks produced by CropFormer.
The decoupled pipeline between CropFomer and GELLA offers several advantages over joint training pipelines for multiple tasks.
First, CropFormer ensures high-quality entity segmentation results, which can then be encoded into the GELLA model for effective grounding.
Second, GELLA remains highly flexible and can process mask inputs from any segmentation framework, not just CropFormer.

Moreover, the design proposed in the GELLA module, including the two single-modal encoders and two dual-modal decoders, is the key component to grounding in the ESG pipeline.
Specifically, the encoders include a lightweight backbone, e.g., MobileNetV2~\cite{sandler2019mobilenetv2}, to process segmentation colormaps and the CLIP vision backbone to handle low-resolution images.
These encoders are computationally efficient due to the usage of lightweight backbones and low-resolution images.
In decoding, the mask decoder integrates visual and colormap features to generate entity visual embeddings for mask reconstruction.
Meanwhile, the large language model's text decoder extracts semantic nouns from the caption and outputs the corresponding text embeddings. 
These visual and text embeddings facilitate the association process in the association module. 
We note that such a design differs significantly from other pooling-based works that can encode global context instead of local features for grounding.

\section{Related Work}

\noindent \textbf{Image Segmentation Dataset.}
Numerous datasets have been proposed for semantic, instance, and panoptic segmentation,~\eg, Microsoft COCO~\cite{lin2014microsoft}, ADE20K~\cite{zhou2017scene}, KITTI~\cite{geiger2012we}, LVIS~\cite{gupta2019lvis}, PASCAL-VOC~\cite{everingham2010pascal}, Open-Images~\cite{kuznetsova2020open},  and Cityscapes~\cite{cordts2016cityscapes}. 
In addition, there are some datasets designed for specific scenarios, such as amodal segmentation (COCO-Amodal~\cite{zhu2017semantic}, KINS~\cite{qi2019amodal}), human segmentation (CelebAMask-HQ~\cite{CelebAMask-HQ}, LIP~\cite{gong2017look}, MHP~\cite{zhao2018understanding}) and domain adaptation (Synscapes~\cite{wrenninge2018synscapes}, and GTA5~\cite{richter2016playing}). 
Despite significant contributions from these datasets, it is of great interest to fulfill the needs of real-world applications with high-quality images with large diversity. 
For example, a segmentation model should be robust to high-resolution images from different domains. 
Meanwhile, models should be able to segment unseen objects in the class-agnostic setting. 
Most relevant to our work is the ADE20K dataset~\cite{zhou2017scene}, which has large-scale open-vocabulary categories. 
%
However, the collected images in ADE20K are of low resolution, like 400px, and from narrowed domains. 
In our dataset, we collect the images from multiple image domains, including indoor, outdoor, street scenes, and even cartoon and remote image domains. 
Over 80\% of the image resolution
falls within the high-resolution range of 2000px and 8000px. 
Compared to the ADE20K~\cite{zhou2017places} and LVIS~\cite{gupta2019lvis} datasets with predefined category lists, our annotation process is different.
We first conduct class-agnostic mask annotation on each entity.
The category information is then labeled in an open-vocabulary manner.

\noindent \textbf{Scene Parsing Methods.}
Convolution-based dense or transformer-based query predictions mainly develop scene parsing methods. 
Methods based on convolution-based dense prediction often use explicit localization information, such as bounding box proposals or pixel positions, for pixel-level mask predictions.
These approaches include
FCN~\cite{long2015fully}, DeepLab~\cite{chen2017deeplab}, PSPNet~\cite{zhao2017pyramid}, Mask R-CNN~\cite{he2017mask}, PANet~\cite{liu2018path}, SOLO~\cite{wang2019solo}, CondInst~\cite{tian2020conditional}, PanopticFPN~\cite{kirillov2019panopticfpn}, and PanopticFCN~\cite{li2022fully}. 
Motivated by the success of transformers DETR~\cite{carion2020end} for detection tasks~\cite{lin2017feature,lin2017focal,dai2017deformable,qi2021multi}, numerous methods, \eg, Max-Deeplab~\cite{wang2021max} and Mask2Former~\cite{cheng2022masked}, directly predict segmentation masks without the guidance of bounding boxes. 
The key to these methods is to use learnable queries to model the implicit location of the potential instance area.
%
Although query-based methods are effective, they cannot benefit from TTA, such as ensembling multi-scale results, which is commonly used to boost the accuracy of convolution-based methods.
Specifically, it is challenging for a query-based method to ensemble multiple results from augmented image views due to the lack of exact locations for queries.
Motivated by video-level Mask2Former, we endow the queries with the ability to connect the same entities among different image views. 
Our approach can enhance high-resolution image segmentation by combining the results from the full image and local crops.

\begin{figure*}[t!]
  \centering
  \includegraphics[width=1.0\linewidth]{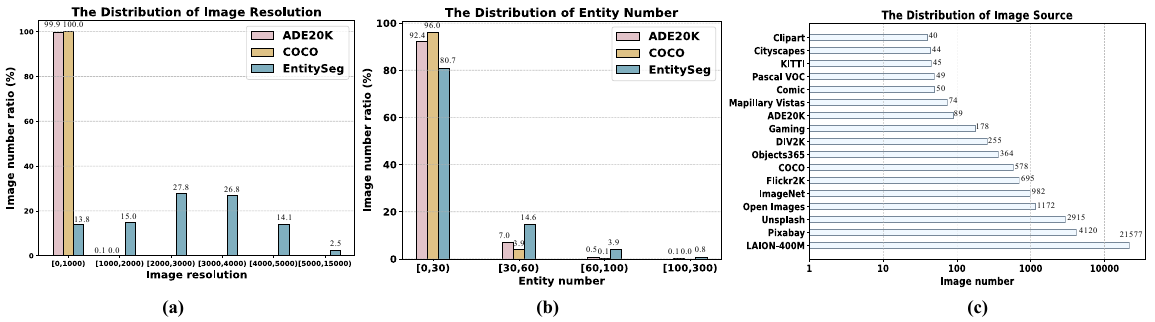}
  \vspace{-0.3in}
  \caption{\textbf{(a), (b): Distributions of image resolutions and average number of entities among ADE20K, COCO, and EntitySeg. (c): Distribution of image sources from which we collected EntitySeg images.}}
  \vspace{-0.1in}
  \label{fig:statistics_01}
\end{figure*}

\vspace{1mm}
\noindent \textbf{Open Image Segmentation.}
%
Another line of image segmentation focuses on the generalization ability to unseen categories or image domains in a class-agnostic manner, as seen in methods like SAM~\cite{kirillov2023segment} or entity segmentation~\cite{qi2021open,qi2023high,qi2023aims}. 
Combined with class-agnostic segmentation results, some open-vocabulary segmentation (OVS)~\cite{li2022language,ghiasi2021open,zhou2022maskclip,wu2023towards,xu2023side,sun2024clip} refers to the task of segmenting images into several regions while assigning to each region a potentially \textit{unlimited or even open set} of the object category. 
For example, Grounded-SAM~\cite{ren2024grounded} and Grounded-HQ-SAM~\cite{ke2024segment} prompt the open-vocabulary box-level proposals generated from Ground-DINO~\cite{liu2025grounding} for mask prediction.
Although these methods exhibit good generalizability, they cannot associate images with natural language prompts. 
To enhance versatility, our aim is to facilitate generalized entity segmentation from complex text prompts using large language models.
Compared to the open-vocabulary segmentation methods, our approach is more adaptable to longer user prompts. 
This is achieved by employing a class-agnostic segmentation followed by open-vocabulary classification.

\vspace{1mm}
\noindent \textbf{Large Multimodal Model.}
Large language models (LLMs) have demonstrated remarkable versatility and capabilities across various tasks.
Built on the strengths of LLMs, large multimodal models (LMMs) aim to integrate multimodal skills to achieve versatility across diverse domains, including language, vision, and other modalities.
Notable contributions from models such as LLaVA~\cite{liu2023llava,liu2023improvedllava}, InstructBLIP~\cite{dai2023instructblip}, and MiniGPT-4~\cite{zhu2023minigpt4} enable the generation of textual responses from both image and text inputs by instruction tuning pre-trained LLMs.
However, these methods are not able to perform region-specific pixel-level visual grounding.
Recently, there have been studies examining grounded text response generation with LMMs. Approaches such as KOSMOS-2~\cite{peng2023kosmos2}, Shikra~\cite{chen2023shikra}, and Ferret~\cite{you2023ferret} encode bounding box coordinates into a sequence of location tokens to enable bounding box generation. 
These approaches rely on language models to generate grounding outputs and cannot perform fine-grained segmentation.
BuboGPT~\cite{zhao2023bubogpt} utilizes an off-the-shelf visual grounding module to explore the fine-grained relation between different visual objects and modalities.
In~\cite{lai2023lisa}, LISA incorporates a segmentation token in the vocabulary, which is then decoded into a segmentation mask for fine-grained reasoning.
Recently, GLaMM~\cite{hanoona2023GLaMM}, PixelLM~\cite{ren2023pixellm}, and OMG-LLAVA~\cite{OMGLLaVA} annotate a new dataset and propose the corresponding strategy for region-level caption and referring segmentation tasks.
Unlike existing approaches that emphasize a unified model for joint training, our method decouples segmentation and grounding into two distinct stages, enabling fine-grained segmentation for multiple generalized entities.

\section{EntitySeg Dataset}
The EntitySeg dataset contains 33,227 images with high-quality mask annotations. 
Compared with existing datasets, there are three distinct properties in EntitySeg. 
%
First, 71.25\% and 86.23\% of the images are high resolution with at least 2000$\times$2000 pixels (we denoted pixels as px for simplicity) and 1000$\times$1000 pixels, which are more consistent with current digital imaging trends.
Second, our dataset is not limited to predefined classes. 
We treat each semantically coherent region in the images as an entity, even 
if it is a blurred region or cannot easily be semantically recognized. 
Third, the mask annotations along the boundaries are more accurate than existing datasets, as shown in Figure~\ref{fig:comp_coco}. 
%

\subsection{Image Collection}
Inspired by the collection criterion of COCO~\cite{lin2014microsoft}, we collect most non-iconic images with more contextual information and non-canonical viewpoints. 
In addition, the image domains should be as diversified as possible to guarantee substantial domain diversity. 
Therefore, our dataset's image sources include several public datasets and the Internet, where the images are permitted for academic research use. 
For public dataset sources, we select part of the images from
COCO~\cite{lin2014microsoft}, ADE20K~\cite{zhou2017scene}, 
Pascal VOC~\cite{everingham2010pascal}, Cityscapes~\cite{cordts2016cityscapes}, 
Mapillary Vistas~\cite{neuhold2017mapillary}, ImageNet~\cite{krizhevsky2017imagenet}, 
Open Images~\cite{kuznetsova2020open}, DIV2K~\cite{agustsson2017ntire}, Flick2K~\cite{wang2019flickr1024}, Clipart~\cite{inoue2018cross},
Comic~\cite{inoue2018cross}, DOTA~\cite{xia2018dota} and some computer game screenshots from Synscapes~\cite{wrenninge2018synscapes} and GTA5~\cite{richter2016playing}. From the Internet, we crawled mainly high-resolution images from Pixabay~\cite{pixabay}, Unsplash~\cite{Unsplash}, and LAION-400M~\cite{schuhmann2021laion}. 
In Figure~\ref{fig:statistics_01}(c), we present the distribution of image sources, highlighting that most images originate from high-resolution sources such as Pixabay, Unsplash, and LAION-400M. 
%
Figure~\ref{fig:statistics_01}(a) shows that our dataset contains a higher proportion of high-resolution images, with resolutions normally distributed from 0px to 15,000px.
%
In contrast, all images in ADE20K~\cite{zhou2017scene} and COCO~\cite{lin2014microsoft} are less than 1000px.

\subsection{Image Annotation}
The annotation process for our dataset mainly consists of three steps. 
For an image, we first annotate all entities with class-agnostic pixel-level masks where each mask does not overlap.
Each mask is annotated with a class label based on a large vocabulary class set, which is not fixed but is continuously updated over time. 
There are two particular considerations here:
1) we annotate the entity as ``\textit{unknown}'' or ``\textit{blurred}'' if it cannot be named easily or is severely blurred; 2) the entity is annotated as a  ``\textit{supercategory}\_other'' if we merely know it is a \textit{supercategory} but unable to identify its fine-grained class.
Our annotation process is the direct opposite of that of popular segmentation datasets such as COCO~\cite{lin2014microsoft} and ADE20K~\cite{zhou2017places}, where class sets are first defined before masks are annotated based on the predefined classes.

Table~\ref{Tab:comp_entity_other} shows that our dataset's mask annotations have a greater proportion of covered image areas and a more significant number of masks on average than COCO- and ADE20K-Panoptic. 
This is because the annotation process allows us to consider all the semantically coherent regions and then assign class labels to them.
In addition, our annotation procedure is more analogous to the human visual system. 
As shown in~\cite{Marr1982Vision}, the human vision system is inherently class-agnostic and can recognize entities without understanding their usage and purpose.

Annotation consistency is crucial to any human-labeled dataset since it implies whether the annotation task is well-defined and proper. 
To study this, we randomly selected 500 images from EntitySeg ($1.5\%$ of the entire dataset), and asked another two annotators to annotate them four months after the first round. 
Table~\ref{tab:consistency}(a) shows that the class-agnostic mask mAP in the first step of the annotation process is compared to that of the other two annotators. 
We use one as ground truth and the other as prediction results. 
Table~\ref{tab:consistency}(b) shows the category consistency by accuracy (ACC) under the same mask annotations. 
These tables demonstrate that our EntitySeg has high annotation consistency in the mask and category labeling stages.

\begin{table}[tp]
\centering
\tiny
\caption{\textbf{Statistical comparison between COCO~\cite{lin2014microsoft}, ADE20K~\cite{zhou2017scene} and EntitySeg.} ImageRes (avg), EntityNum (avg), Valid Area, Entity Complexity, and Entity Simplicity refer to the average value of image resolution size, entity numbers per image, valid area ratio per image, average entity complexity, and simplicity, respectively. EntityNum (max) means the maximum per-image number of entities within each dataset.}
\label{Tab:comp_entity_other}
\begin{tabular}{ccccccc}
\cellcolor{lightgray!30} Dataset & \cellcolor{lightgray!30} \makecell{ImageRes\\(avg)$\uparrow$} & \cellcolor{lightgray!30} \makecell{EntityNum\\(avg)$\uparrow$} & \cellcolor{lightgray!30} \makecell{EntityNum\\(max)$\uparrow$} & \cellcolor{lightgray!30} \makecell{Entity\\Complexity$\downarrow$} &
\cellcolor{lightgray!30} \makecell{Entity\\Simplicity$\downarrow$} & \cellcolor{lightgray!30} \makecell{Valid\\Area$\uparrow$}  \\ \toprule
COCO & 522.5 & 11.2 & 95 & 0.758 & 0.581 & 0.891 \\
ADE20K & 461.3& 13.6 & \textbf{255} & 0.802 & 0.606 & 0.914 \\
EntitySeg & \textbf{2700.7} & \textbf{18.1} & 236 & \textbf{0.719} & \textbf{0.538} & \textbf{0.999} \\ \bottomrule
\end{tabular}
\end{table}

\subsection{Dataset Statistics}

The EntitySeg dataset annotates all images at the entity level, regardless of whether they belong to things or stuff as labeled in other datasets. 
As shown in Figure~\ref{fig:comp_coco}, we show some low- and high-resolution photos and their mask annotations to highlight our high-quality pixel-level annotation quality.
For example, one can zoom in on the Unsplash image to see how fine-grained the railing's mask is.

Table~\ref{Tab:comp_entity_other} shows the quantitative comparisons among our dataset, COCO-Panoptic~\cite{lin2014microsoft}, and ADE20K-Panoptic~\cite{zhou2017scene}.
In the EntitySeg dataset, each image has 18.1 entities on average, which is more than 11.2 and 13.6 entities in COCO and ADE20K. 
The detailed comparisons among the three datasets regarding the distribution of entity numbers is illustrated in Figure~\ref{fig:statistics_01}(b). 
Furthermore, the shapes of entities in our dataset are more complex than those in COCO~\cite{lin2014microsoft} and ADE20K~\cite{zhou2017scene}, as indicated by the columns labeled Entity Complexity and Entity Simplicity in Table~\ref{Tab:comp_entity_other}.
%
More details regarding the shape convexity and simplicity of an entity mask are presented in the appendix.

\vspace{1mm}
\noindent \textbf{Class-Aware Annotation.}
We select a subset of 11,580 images from the entire dataset for class labeling, which forms the EntityClass dataset. 
In our annotation process, the class is labeled in an open-vocabulary and free-form manner, then the class set is continually expanded.
We organize and merge the annotated classes
according to Word-Tree~\cite{ILSVRC15}.
This results in 535 things and 109 stuff classes for EntityClass that follow Zipf's law. 
More details regarding the entity and pixel-level class frequency distribution are presented in the appendix.

\begin{table}[t!]
\caption{\textbf{Annotation consistency of class-agnostic localization and class-aware categories among annotators.}}
\label{tab:consistency}
\begin{minipage}{\textwidth}
\begin{minipage}[t]{0.22\textwidth}
            \centering
            \footnotesize
            \setlength{\tabcolsep}{2pt}
            \begin{tabular}{cccc}
             \cellcolor{lightgray!30} & \cellcolor{lightgray!30} ann$_{1}$ & \cellcolor{lightgray!30} ann$_{2}$ & \cellcolor{lightgray!30} ann$_{3}$\\ \toprule
            \cellcolor{lightgray!30} ann$_{1}$ & - & - & -\\ 
            \cellcolor{lightgray!30} ann$_{2}$ & 90.6 & - & -\\ 
            \cellcolor{lightgray!30} ann$_{3}$ & 91.2 & 92.1 & -\\ \bottomrule
            \end{tabular}
        class-agnostic
        \end{minipage}
        \begin{minipage}[t]{0.18\textwidth}
        \centering
        \footnotesize
        \setlength{\tabcolsep}{2pt}
        \begin{tabular}{cccc}
             \cellcolor{lightgray!30} & \cellcolor{lightgray!30} ann$_{1}$ & \cellcolor{lightgray!30} ann$_{2}$ & \cellcolor{lightgray!30} ann$_{3}$\\ \toprule
             \cellcolor{lightgray!30} ann$_{1}$ & - & - & -\\
             \cellcolor{lightgray!30} ann$_{2}$ & 95.4 & - & -\\ 
             \cellcolor{lightgray!30} ann$_{3}$ & 94.8 & 95.2 & -\\ \bottomrule
        \end{tabular}
        class-aware 
        \end{minipage}
\end{minipage}
\end{table}

\section{The ESG Pipeline}
Given an image $\mathbf{I}^{o}\in\mathbb{R}^{1\times H\times W\times 3}$, where $H$ and $W$ denote the image height and width, the entity segmentation and grounding (ESG) pipeline aims to generate $N^{e}$ entity masks $\mathbf{U}^{m}_{e}\in\{0,1\}^{N^{e}\times H\times W}$, an image-level caption, and the correspondence between the predicted entity masks and the semantic nouns appearing in the caption. Here, the image-level caption can be automatically generated or directly provided by the user via prompting.

As shown in Figure~\ref{fig:gella_overview}, the proposed ESG pipeline mainly consists of two core modules: CropFormer for class-agnostic entity-level segmentation and GELLA for entity grounding. This design is inspired by the standard two-stage detection paradigm (e.g., the R-CNN series), where class-agnostic proposals are generated first and then used for subsequent semantic understanding.

\subsection{CropFormer for Segmentation}
\label{sec:cropformer}
We first introduce the high-quality segmentation part of our ESG pipeline.
Motivated by recent query-based segmentation methods~\cite{cheng2021per,cheng2022masked,wang2021max,yu2022k}, we propose CropFormer, a transformer-based set-prediction segmentation method capable of fusing mask predictions from multiple image views.
Given multiple views of an image $\mathbf{I} \in \mathbb{R}^{T \times H \times W \times 3}$ where $T$, $H$ and $W$ correspond to the view number, height, and width of the image, the CropFormer targets to learn $\mathbf{Q}\in \mathbb{R}^{N\times K}$ to generate mask embeddings $\mathbf{E} \in \mathbb{R}^{N \times 1 \times 1 \times 1 \times K}$. 
$N$ is the number of queries with dimension $K$.
Then $\mathbf{E}$ operates as a set of convolution filters on the features of the mask $\mathbf{P_2} \in \mathbb{R}^{T \times H' \times W'}$ to generate $N$ segmentation masks $\mathbf{U^m} \in \mathbb{R}^{N \times T \times H \times W}$ and entityness score $\mathbf{U^b} \in \mathbb{R}^{N}$. $H'$ and $W'$ are the height and width of the down-sampled features.
Finally, we select \(N^{e}\) entity masks \(\mathbf{U}^{m}_{e}\) by applying a fixed threshold to the entityness scores \(\mathbf{U}^{b}\) and then merge the \(T\) views.

\begin{figure*}[t!]
  \centering
   \includegraphics[width=1.0\linewidth]{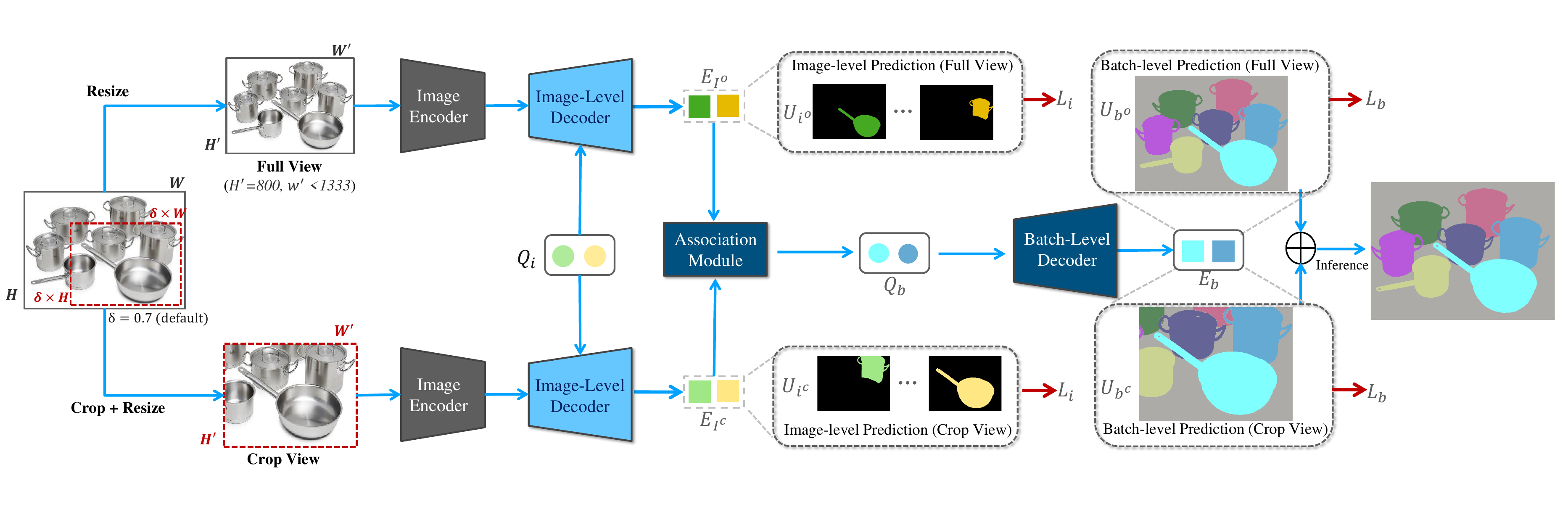}
   \caption{\textbf{Framework of the proposed CropFormer.} The red box indicates a cropped region randomly sampled from four fixed image corners. In image-level prediction, the same entity across different image views may be assigned to different queries. Our association module and batch decoder can effectively associate the same entity across different views with a single query.}
   \label{fig:framework}
\end{figure*}

However, merging mask results of \(T\) views is not trivial.
%
We observe that the same queries are not robust in representing the same entities across different image crops.
Consequently, we cannot leverage these queries to ensemble inference results for the same entities across multiple crops. 
To address this limitation, we introduce a novel association module and a batch-level decoder to achieve the goal of CropFormer, which takes advantage of global context from the full image and fine-grained local details from crops for high-quality segmentation. 
The framework of CropFormer is illustrated in Figure~\ref{fig:framework}.

\noindent \textbf{Crop Dataloader.} In the CropFormer framework, the crop dataloader is designed to generate a batch of images that simultaneously include the full images and their corresponding crops. 
There are two steps in our dataloader: \textit{crop} and \textit{resize}. 
First, we augment the full image $I^o$ with a crop that is part of the full image. The crop is randomly extracted from one of the fixed image corners: upper-left, upper-right, bottom-left, and bottom-right. 
The crop size is controlled by a fixed ratio hyperparameter $\delta \in \mathbb{R}$ relative to the full image size. 
Then, we resize both the full image and crop to the same size. In this way, $T$ is 2 in the input tensor $\mathbf{I}$. Compared to the full image, our corner crops preserve more image details (\eg, local information) that are useful for high-quality segmentation.
With this dataloader $\mathcal{D}$, we have: 
\begin{equation}
\mathbf{I} = \{I^o, I^c \}, \ \ I^o \sim \mathcal{D}
\end{equation}
where $I^o$ is sampled from $\mathcal{D}$, and $I^c$ is the cropped image controlled by fixed ratio $\delta$. 
Unlike the random cropping in~\cite{seqformer,heo2022vita}, we adopt a fixed cropping solution to keep the training and inference consistent. 
This rigid cropping strategy provides better inference efficiency since, rather than bunches of random crops, evaluating a fixed number of crops during inference is sufficient.

\vspace{1mm}
\noindent \textbf{Image Encoder and Decoder.} 
Based on learnable image-level queries $\mathbf{Q}_{i} \in \mathbb{R}^{N\times K}$,  we use an image encoder $\Theta$ and an image-level decoder $\Phi_{i}$ to generate image-level embeddings $\mathbf{E_i} \in \mathbb{R}^{N \times 2 \times 1 \times 1 \times K}$ for the full input image and its crop. 
Given the input tensor ($\mathbf{I}$) and queries ($\mathbf{Q}_{i}$), we define $\mathbf{E_i}$:
\begin{equation}
\mathbf{E_i} = \Phi_i(\mathbf{Q_i}, \Theta(\mathbf{I})),
\end{equation}
where $\Phi_{i}(\cdot)$ is a transformer-based image-level decoder.
$\mathbf{E_i}$ is used for image-level entityness prediction and pixel-level prediction using the low-level image feature 
$\mathbf{P_2}$ (derived from the image encoder).
The prediction process is formulated as:
\begin{equation}
\mathbf{U^e_i}, \mathbf{U^m_i} = \text{PredHead}_{\mathbf{i}}(\mathbf{E_i}, 
\mathbf{P_2})\label{eq:eq3},
\end{equation}
where $\mathbf{U^e_i}$ and $\mathbf{U^m_i}$ denote
the entityness prediction and pixel-level mask outputs. 
We use the subscript $i$ to differentiate image-level embeddings and mask outputs from those of the association module.

\vspace{1mm}
\noindent \textbf{View Association Module.}
Similar to~\cite{cheng2021mask2former}, our view association module aims to generate batch queries $\mathbf{Q_b}$ that are shared by the full image and its crop to represent the same entities consistently. 
However, instead of treating $\mathbf{Q_b}$ as learnable model parameters, we generate $\mathbf{Q_b}$ directly from $\mathbf{E_i}=\{\mathbf{E}_{I^o}, \mathbf{E}_{I^c}\}$ since 
$\mathbf{E_i}$ already contains strong segmentation features.
Specifically, we use a transformer architecture with cross-attention ($f_{\text{XAtt}}$) and self-attention ($f_{\text{SAtt}}$) to obtain $\mathbf{Q_b}$:
\begin{equation}
\mathbf{Q_{b}} = \text{FFN}(f_{\text{SAtt}}(f_{\text{XAtt}}(\underbrace{f_\text{q}(\mathbf{E}_{I^o})}_\text{query}, \underbrace{f_\text{k}(\mathbf{E_i})}_\text{key}, \underbrace{f_\text{v}(\mathbf{E_i}}_\text{value})))),
\end{equation}
where FFN is a feed-forward network,
$f_{\{\text{q},\text{k},\text{v}\}}(\cdot)$ are linear transformations.
Considering the importance of full image for entity segmentation, 
we take the image-level embeddings of full image $\mathbf{E}_{I^o}$ as \textit{query}, while treating all image-level embeddings $\mathbf{E_i}$ as \textit{key} and \textit{value} in a transformer. 

\begin{figure}[t!]
  \centering
   \includegraphics[height=1.3in, width=0.95\linewidth]{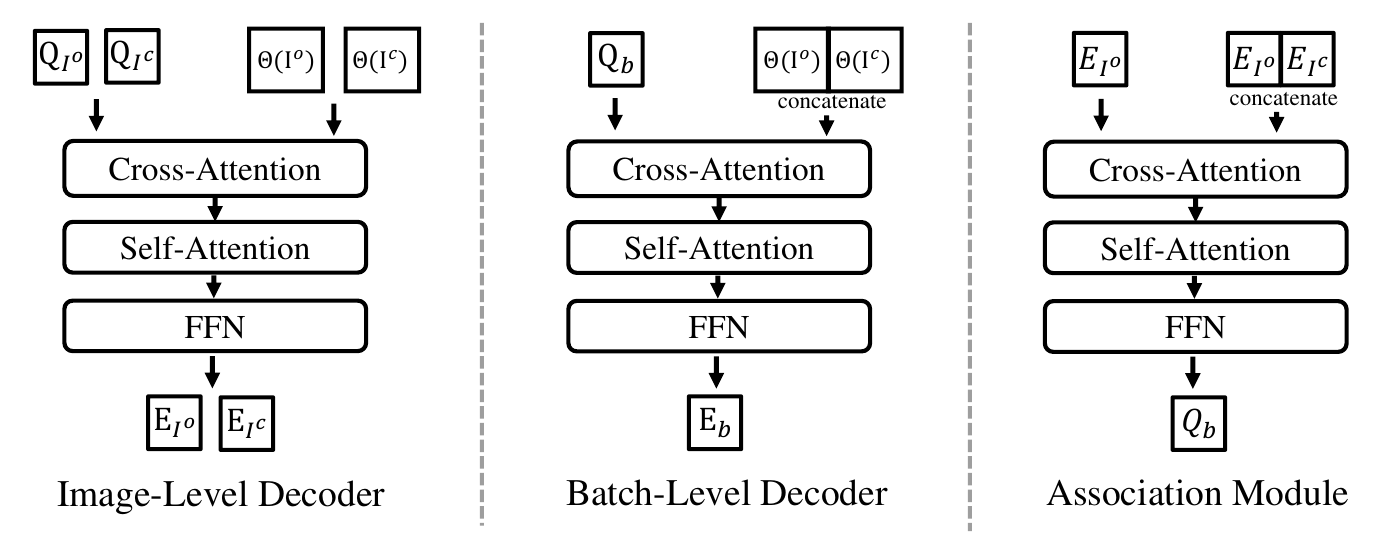}
   \vspace{-0.1in}
   \caption{\textbf{Illustration of image- and batch-level decoder and association module.}}
   \vspace{-0.2in}
   \label{fig:com_decoder}
\end{figure}

\vspace{1mm}
\noindent \textbf{Batch-Level Decoder.}
Given $\mathbf{Q_b} \in \mathbb{R}^{N \times 1 \times 1 \times 1 \times K}$, we obtain batch embeddings $\mathbf{E_b} \in \mathbb{R}^{N \times 1 \times 1 \times 1 \times K}$:
\begin{equation}
\mathbf{E_b} = \Phi_b(\mathbf{Q_b}, \Theta(\mathbf{I})),
\end{equation}
where $\Phi_b(\cdot)$ denotes the batch-level decoder. 
%
Thus, we share $\mathbf{E_b}$ to the shape of $N \times 2 \times 1 \times 1 \times K$ between the entire image and its crop.
Finally, the broadcasted $\mathbf{E_b}$ and low-level batch features $\mathbf{P_{2}}$ are used for batch-level mask classification and pixel-wise prediction:
\begin{equation}
\mathbf{U^e_b}, \mathbf{U^m_b} = \text{PredHead}_{\mathbf{b}}(\mathbf{E_b}, 
\mathbf{P_{2}}).
\end{equation}
For the image-level decoder, the queries are separate for each image/crop. 
The queries are shared between the image and its crop for the batch-level decoder. 
We illustrate the differences between the image- and batch-level decoders in Figure~\ref{fig:com_decoder}, where we regard two image views as an entirety.

\begin{figure*}[th!]
\centering
\includegraphics[width=\textwidth]{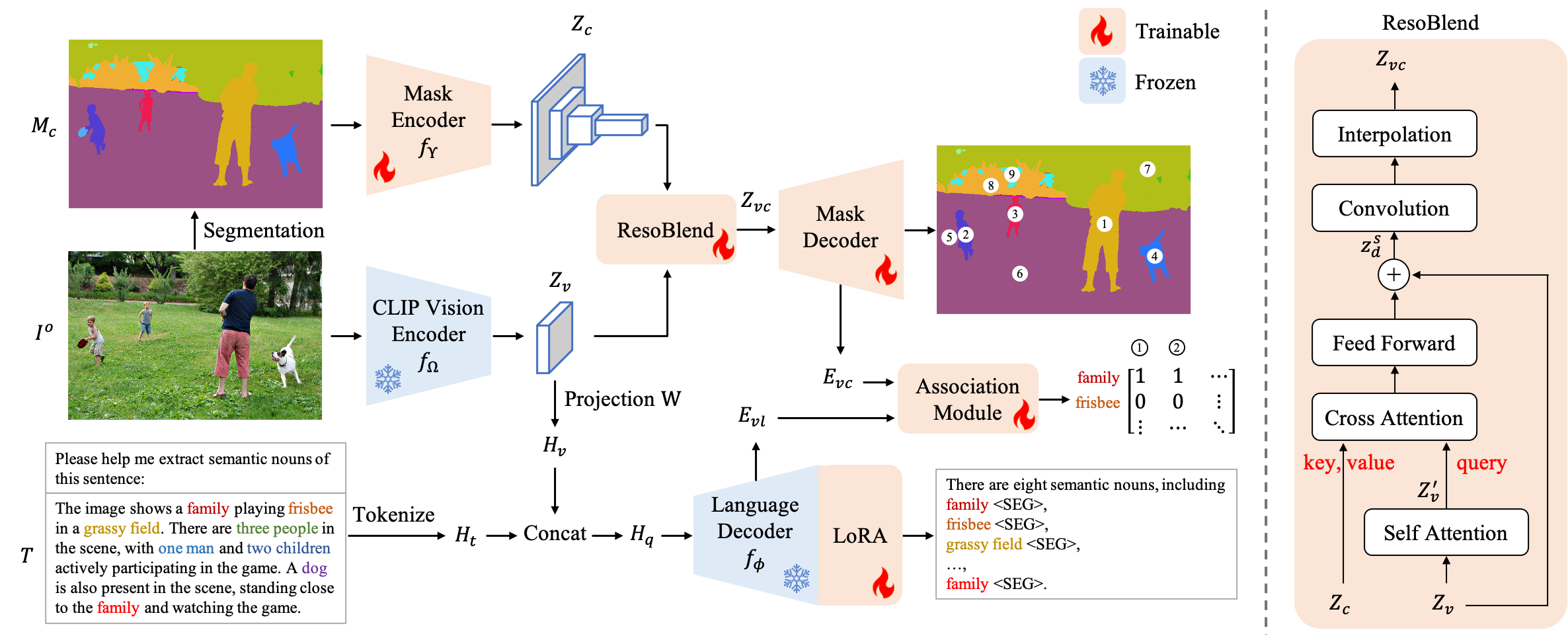}
\vspace{-4mm}
\caption{\textbf{Left: Overview of the GELLA framework.} Given an image $\mathbf{I^o}$ and its corresponding caption $T$, we aim to generate a panoptic segmentation that densely grounds the semantic nouns in the caption.
We first obtain an entity-level mask colormap $\mathbf{M_c}$ using a class-agnostic segmentation model.
The mask and image are encoded by a mask encoder $f_{\Upsilon}$ and a CLIP vision encoder $f_{\Omega}$, respectively. 
The two extracted features are then fused by the ResoBlend module and fed into a mask decoder to reconstruct the mask. 
For the language part, we prepend an instruction to the caption $T$ for extracting semantic nouns and proceed with the tokenizer. 
The visual tokens $\mathbf{H_v}$ and textual tokens $\mathbf{H_t}$ are concatenated and fed into the language decoder $f_{\phi}$ to generate $<$SEG$>$ tokens as features of each semantic noun. 
The association module then computes the similarity between the embeddings of semantic nouns and visual entities. \textbf{Right: Illustration of the ResoBlend Module.}}
\label{fig.framework}
\end{figure*}

\vspace{1mm}
\noindent \textbf{Training.} 
During training, we use two separate losses $\mathcal{L}_{i}$ and $\mathcal{L}_{b}$ for image- and batch-level predictions, concerning ground truth $\mathbf{G}$.
The main difference between the two losses lies in
whether the same entities in the full image and the crop are tied to the same queries or not.

The overall training loss for our model is defined as:
\begin{align}
\mathcal{L}_{\text{CropFormer}} = & \sum_{\mathbf{k}\in \{\mathbf{i}, \mathbf{b}\}} \mathcal{L}_{\mathbf{k}}^{\text{ce}}(\mathbf{U^e_k}, \mathbf{G^e_k}) + \sum_{{\mathbf{k}}\in \{\mathbf{i}, \mathbf{b}\}} \mathcal{L}_{\mathbf{k}}^{\text{bce}}(\mathbf{U^m_k}, \mathbf{G^m_k}) + \nonumber\\
& \sum_{\mathbf{k}\in \{\mathbf{i}, \mathbf{b}\}} \mathcal{L}_{\mathbf{k}}^{\text{dice}}(\mathbf{U^m_k}, \mathbf{G^m_k}),
\end{align}
where $\mathcal{L}_{\mathbf{i}}^{\text{ce}}$ and $\mathcal{L}_{\mathbf{b}}^{\text{ce}}$ denote binary cross-entropy loss for image- and batch-level entityness prediction. 
Similarly, $\mathcal{L}_{\mathbf{i}}^{\text{bce}}$, $\mathcal{L}_{\mathbf{b}}^{\text{bce}}$, $\mathcal{L}_{\mathbf{i}}^{\text{dice}}$ and $\mathcal{L}_{\mathbf{b}}^{\text{dice}}$ denote the binary cross-entropy and dice loss for image- and batch-level mask prediction.

\vspace{1mm}
\noindent \textbf{Inference.}
In CropFormer, the same entities between the full image and its crop are represented by the same queries for the association module. 
We average the per-pixel mask predictions obtained from the full image and every crop of the four corners for the segmentation output. 
The confidence score of each entity comes from the batch-level entityness prediction score.

\subsection{GELLA for Entity Grounding}
In this subsection, we introduce GELLA, the grounding part in our decoupled pipeline.
Here, we only use the full image view $\mathbf{I^o}$ together with high-quality entity segmentation results $\mathbf{U}^{m}_{e} \in \mathbb{R}^{N^e \times H\times W}$ using CropFormer and a long caption $\mathbf{J}$ as our inputs.
%
Our GELLA framework then parses the caption $\mathbf{J}$ to $v$ semantic nouns $\mathbf{X} \in \{x_0,...,x_v\}$ and assigns them to each mask of $\mathbf{U}^{m}_{e}$. 
We note that the assignment target is a one-to-many association because some semantic nouns, such as `three persons' may correspond to several masks.


As illustrated in Figure~\ref{fig.framework}, the GELLA framework has two single-modal encoders and two dual-modal decoders. 
First, the single-modal encoders process the image and the segmentation results to generate the corresponding features. 
Then, these features are fused to create dual-modal features, integrating image/colormap and image/language information, respectively.
%
%
For decoding, the mask decoder integrates these fused features to generate entity visual embeddings for mask reconstruction.
Meanwhile, the large language model's text decoder extracts semantic nouns from the text prompt and outputs the corresponding text embeddings.
These visual and text embeddings facilitate the association process in the association module.

Specifically, the encoders include a lightweight backbone, such as MobileNetV2, for processing segmentation colormaps and the CLIP vision backbone for handling low-resolution images. 
These encoders are computationally efficient due to the usage of lightweight backbones and low-resolution images.

For the mask decoder, mask reconstruction facilitates the GELLA model to derive clues from features extracted by the colormap encoder. 
This approach reduces the processing burden on the CLIP vision encoder, which would otherwise require high-resolution images. 
Unlike cropping mask areas of entities for open-vocabulary classification, the GELLA framework offers more contextual information, thereby effectively addressing nouns with multiple entities or specific relationships. 
For example, it can distinguish between phrases like `the three dogs in the grass' and `one dog near the door' in one caption. 
The ablation studies in Table~\ref{tab:abl_reason_open} on reasoning segmentation and open-vocabulary segmentation prove the effectiveness of the GELLA framework.

\vspace{1mm}
\begin{table*}[t!]
\centering
\vspace{-2mm}
\caption{\textbf{One example to illustrate our instruction-based training data.} We prompt GPT-4 for image description generation and semantic noun extraction.}
\vspace{-2mm}
\label{tab:example_instruction}
\begin{minipage}{0.99\textwidth}
\centering
\begin{tcolorbox} 
\centering
\footnotesize
\begin{tabular}{p{\textwidth} c}
\VarSty{{\bf User Type 1: Image Description Generation}} & \\
{\bf User:}
$<$IMAGE$>$ Please help me describe the image.
& \hspace{-4.5cm} \multirow{4}{*}{ \includegraphics[height=2.2cm]{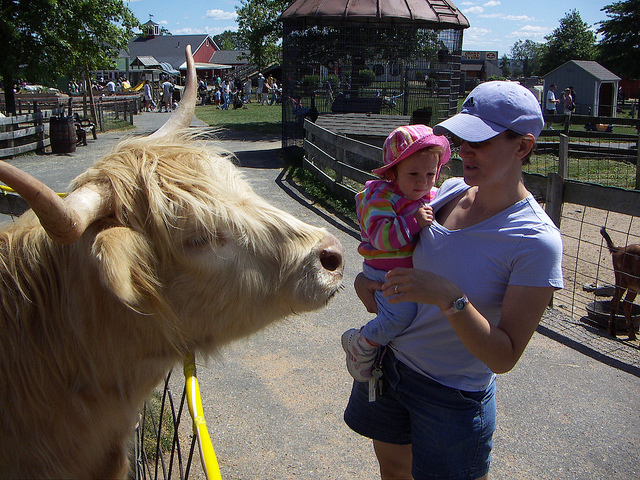} }
\\
{\bf Assistant:}
This image depicts an outdoor scene at what appears to be a petting zoo or farm. On the left, here's a light \\brown cow with a shaggy coat and long horns peering over a yellow barrier, possibly looking for food or attention \\from the visitors. To the right, a woman wearing a cap and sunglasses is holding a baby. The baby is wearing a pink \\hat with a striped pattern and looks curiously at the cow. In the background, more animal enclosures and several \\visitors are enjoying the sunny day. The setting suggests a family-friendly environment where people can interact with \\farm animals. \\

\VarSty{ {\bf User Type 2: Semantic Noun Extraction} } & \\
{\bf User:}
$<$IMAGE$>$ Please help me extract each semantic noun of the sentence. The sentence is $<$EXPRESSION$>$. 
 
{\bf Assistant:}
There are twenty one semantic nouns, including petting zoo $<$SEG$>$, light brown cow $<$SEG$>$, shaggy coat $<$SEG$>$, horns $<$SEG$>$, barrier $<$SEG$>$, food $<$SEG$>$, visitors $<$SEG$>$, woman $<$SEG$>$, sunglasses $<$SEG$>$, baby $<$SEG$>$... and animals $<$SEG$>$.
\end{tabular}
\end{tcolorbox}
\end{minipage}
\end{table*}

\vspace{1mm}
\noindent Colormap Encoder. We encode the entity-level binary segmentation masks $\mathbf{U}^{m}_{e}$ into a colormap $\mathbf{M_c} \in [0, 255]^{H\times W\times 3}$ using a colormap encoder:
$\mathbf{M_c} = f_{\Psi}(\mathbf{U^{m}_{e}})$,
where $f_{\Psi}$ indicates the random color assignment for each entity. 
Other assignment methods, such as the location-aware assignment used in Painter~\cite{wang2023images}, do not result in performance gains (see Section~\ref{sec:ablation}).

Similar to the conventional image encoding process, we encode a colormap to the pyramid mask features as
\begin{equation}
\mathbf{Z_c} = f_{\Upsilon}(\mathbf{M_c}),
\end{equation}
where $\mathbf{Z_c} \in \{z^s\}$, with $s \in \{2,3,4,5\}$ is the index of pyramid feature, $z^s \in \mathcal{R}^{\frac{h}{2^s}\times \frac{w}{2^s}\times c^s}$, and $c^s$ is the channel dimension of $z^s$. 

For $f_{\Upsilon}$, we find that existing lightweight structures, such as MobileNet~\cite{sandler2019mobilenetv2} and Swin-Tiny~\cite{liu2021swin}, designed for image classification, can effectively encode the colormap.
%
This is because the entities represented by different colors have explicit boundaries, which provide our framework with a robust shape priority. 
This design eliminates the need of high-resolution images for pixel-level prediction, thereby allowing the usage of a CLIP~\cite{radford2021learning} visual backbone with low-resolution images.

\vspace{1mm}
\noindent \textbf{Image Encoder.} We use a CLIP vision encoder in the ViT structure. 
%
Similar to prior work, our image input maintains its original pre-trained resolution (\eg, 224$\times$224 and 336$\times$336),
\begin{equation}
\mathbf{Z_v} = f_{\Omega}(\text{RESIZE}(\mathbf{I^o})),
\end{equation}
where $\mathbf{Z_v} \in \mathcal{R}^{(\frac{h}{64}\times \frac{w}{64})\times C_v}$ is the grid features after the last transformer layer, and $C_v$ is the dimension of visual features.
The $h \in \{224, 336\}$ and $w \in \{224, 336\}$ are the height and width of resized $\mathbf{I}^{o}$.
Such a design maintains low computation costs, considering the low-resolution image input.
Here, $\mathbf{Z_v}$ can be combined with other modal features $\mathbf{Z_c}$ or language token embedding to create multimodal features $\mathbf{Z_{vc}}$ (visual and mask features) or $\mathbf{Z_{vl}}$ (visual and language features).

\noindent \textbf{Language Decoder.} 
Motivated by the LLaVA~\cite{liu2023llava} design, we use a single linear layer to embed image patch features into the word embedding space.
Specifically, a trainable projection matrix $W$ is used to convert $\mathbf{Z_v}$ into the language embedding space as:
\begin{equation}
\mathbf{H_v} = W \cdot \mathbf{Z_v}.
\end{equation}
We prepend the instruction ``Please help me extract semantic nouns of this sentence:'' to the caption $T$ and proceed with the tokenizer. 
Then, we concatenate the visual tokens $\mathbf{H_v}$ with the textual tokens $\mathbf{H_t}$ to form multi-modal tokens $\mathbf{H_q}$.

We choose LLaMA~\cite{touvron2023llama} as the language decoder $f_{\phi}$, which has been demonstrated to be effective in several multimodal models. 
Here, $\phi$ represents the parameters of LLaMA. The language features are extracted as:
\begin{equation}
\mathbf{Z_{vl}} = f_{\phi}(\mathbf{H_q}).
\end{equation}
$\mathbf{Z_{vl}}$ is processed with a fully connected layer to predict the output sentence's tokens $\mathbf{U^t}$, which should include $<$SEG$>$ tokens for the semantic nouns.
We extract the last-layer embedding corresponding to the $<$SEG$>$ tokens as $\mathbf{E_{vl}}$.
The language loss is defined as:
\begin{equation}
\mathcal{L}_{\text{ans}} = \mathcal{L^{\text{ce}}}(\mathbf{U^t}, \mathbf{G^t}),
\end{equation}
where $\mathcal{L}_{ans}$ is the auto-regressive cross-entropy loss for text generation. $\mathbf{G^t}$ is the ground truth of the text.

We do not use original word embeddings to represent semantic nouns due to their lack of visual context. As a result, original word embeddings cannot distinguish the same word in different contexts, such as `dog on the left' and `dog near the door'.


\vspace{1mm}
\noindent \textbf{Mask Decoder.} We adopt the Mask2Former decoder design~\cite{cheng2022masked} without modifications, which includes the pixel and transformer decoders. 
The mask decoder outputs are presented in a class-agnostic entity segmentation manner. 
Specifically, we obtain a group of embeddings $\mathbf{E_{vc}}\in \mathcal{R}^{q\times 256}$ to generate entities' prediction $\mathbf{U^{ge}}\in \mathcal{R}^{q\times 2}$ and pixel-level mask output $\mathbf{U^{gm}}\in \mathcal{R}^{q\times \frac{h}{4}\times \frac{w}{4}}$. $q$ is the query number designed in the mask decoder. 
The loss for this part is
\begin{equation}
\resizebox{.9\hsize}{!}{$\mathcal{L_{\text{seg}}} = \mathcal{L^{\text{bce}}}(\mathbf{U^{ge}}, \mathbf{G^{ge}}) + \mathcal{L^{\text{bce}}}(\mathbf{U^{gm}}, \mathbf{G^{gm}}) + \mathcal{L^{\text{dice}}}(\mathbf{U^{gm}}, \mathbf{G^{gm}})$},
\end{equation}
where $\mathcal{L}^{\text{bce}}$ and $\mathcal{L}^{\text{dice}}$ are the binary cross-entropy and dice loss. The first term is for entityness prediction, while the second and third terms are for mask prediction. The ground truth for each subtask is $\mathbf{G^{ge}}$ and $\mathbf{G^{gm}}$.
%

We note that the mask decoder is retained to ensure that the framework can reconstruct masks by encoding the color map and outputting the query embeddings $\mathbf{E_{vc}}$ for further association.

\noindent \textbf{ResoBlend Module} is designed to augment pyramid mask features $\mathbf{Z_c}$ with the single-layer image feature $\mathbf{Z_v}$.
Despite being a single-layer image feature, $\mathbf{Z_v}$ contains both low- and high-level image information, as demonstrated by ViTDet~\cite{li2022exploring}. 
The right part of Figure~\ref{fig.framework} illustrates the fusion process in our ResoBlend module using several standard attention modules. 
Specifically, we employ a self-attention block to process $\mathbf{Z_v}$ into $\mathbf{Z'_v}$. 
Subsequently, $\mathbf{Z'_v}$ and the mask feature $z^s$ are fused through cross-attention $f_{\text{XAtt}}$, with $\mathbf{Z'_v}$ as the query and $z^s$ as the key and value:
\begin{equation}
z^s_d = (\text{FFN}(f_{\text{XAtt}}(\underbrace{f_\text{q}(\mathbf{Z'_v})}_\text{query}, \underbrace{f_\text{k}(z^s)}_\text{key}, \underbrace{f_\text{v}(z^s}_\text{value})))) + \mathbf{Z_v}.
\end{equation}
Finally, we use a simple convolution with a kernel size of 3 and a bilinear interpolation operation to transform $z_d^s$ into $z_w^s$:
\begin{equation}
z_w^s = \text{BIINTERP}(\text{CONV}(z^s_d)).
\end{equation}
Thus, we feed $\mathbf{Z_{vc}} \in \{z_w^s\}$ to the mask decoder for further decoding.

\noindent \textbf{Association Module}
We introduce a simple association module to learn cross-modal associations between $<$SEG$>$ token embeddings $\mathbf{E_{vl}}$ from the language decoder and entity embeddings $\mathbf{E_{vc}}$ from the mask decoder. 
The association loss $\mathcal{L_{\text{asso}}}$ is defined as:
\begin{equation}
\mathcal{L_{\text{asso}}} = \mathcal{L^{\text{bce}}}(\text{FC}(\mathbf{E_{vl}})
\boldsymbol{\cdot}\text{FC}(\mathbf{E_{vc}})^T, \mathbf{G^{A}}),
\end{equation}
where FC indicates a \textit{fully-connected} layer, $\boldsymbol{\cdot}$ denotes dot product, and $\mathbf{G^{A}} \in \{0,1\}^{v\times q}$ are binary association ground truth of semantic nouns and entities. 

The overall training loss for GELLA is
\begin{equation}
\mathcal{L_{\text{GELLA}}} = \mathcal{L_{\text{seg}}} + \mathcal{L_{\text{asso}}} + \mathcal{L_{\text{ans}}},
\end{equation}
where we use the uniform weights for these losses as in LLaVA~\cite{liu2023llava,liu2023improvedllava} and Mask2Former~\cite{cheng2022masked}.

\subsection{Data Preparation}
Table~\ref{tab:example_instruction} shows an example of visual instructions for training. Specifically, two types of user instructions are considered: image description and semantic noun extraction.
The GELLA model design allows image descriptions from the output itself instead of relying on user instructions. 
When sampling the image description task, we train only the language part without the segmentation output. 
We prompt GPT-4~\cite{openai2023gpt4} and curate a question list for image description and semantic noun extraction, as shown in Table~\ref{tab:example_instruction}. 
For the association between semantic nouns and entities, we directly use the COCO Panoptic Narrative Grounding dataset~\cite{gonzalez2021panoptic}.

\section{Experiments}
We conduct experiments on three aspects. 
%
First, we benchmark state-of-the-art approaches on the EntitySeg dataset in segmentation tasks. 
Second, we perform ablation studies on CropFormer to prove its effectiveness on high-quality entity segmentation. 
Third, we analyze GELLA for systematic analysis of three tasks: panoptic narrative grounding, referring, and panoptic segmentation.
We note that in Sec.~\ref{sec:cropformer}, we omit the conventional \textit{data batch dimension} to improve clarity, but in practice, we train CropFormer with a batch of multiple dataset images and their crops, following existing methods.


\subsection{EntitySeg Dataset Benchmarks}
\begin{table}[t!]
\centering
\scriptsize
\caption{\textbf{Benchmark on segmentation tasks using the EntitySeg dataset.} ``SEM'', ``INS'', ``PAN'', and ``ENT'' indicate the task of semantic, instance, panoptic, and entity segmentation. 
We use the same hyperparameters as the panoptic segmentation training on the COCO dataset (batch size is 16, and AdamW optimizer with a learning rate of 0.0002), except for the training iterations and learning rate steps, considering the dataset size difference. 
For training entity segmentation on the EntitySeg dataset, we adopt a training schedule of 103,125 iterations and decay the learning rate after 91,575 and 99,414 iterations. 
Given that the class-aware tasks have only one-third of the image numbers of the EntitySeg dataset, we scale down to one-third of the training iterations.}
\label{Tab:aba_benchmark_merge}
\begin{tabular}{cccccc}
\cellcolor{lightgray!30} Task & 
\cellcolor{lightgray!30} Metrics & 
\cellcolor{lightgray!30} Method & 
\cellcolor{lightgray!30} Backbone & 
\cellcolor{lightgray!30} Pretrain & 
\cellcolor{lightgray!30} Performance \\ \toprule
\multirow{5}*{SEM} & \multirow{5}*{mIoU} & Deeplabv3 & R-50 & ImageNet & 27.9 \\ 
& & \multirow{4}*{Mask2Former} & R-50 & ImageNet & 37.8 \\ 
& & & R-50 & COCO & 43.3 \\
& & & Swin-T & COCO & 45.0 \\
& & & Swin-L & COCO & 50.5 \\ \midrule
\multirow{5}*{INS} & \multirow{5}*{AP} & Mask-RCNN & R-50 & ImageNet & 5.0 \\ 
& & \multirow{4}*{Mask2Former} & R-50 & ImageNet & 13.0 \\
& & & R-50 & COCO & 20.3 \\
& & & Swin-T & COCO & 22.7 \\
& & & Swin-L & COCO & 30.3 \\ \midrule
\multirow{5}*{PAN} & \multirow{5}*{PQ} & PanopticFPN & R-50 & ImageNet & 3.6 \\ 
& & \multirow{4}*{Mask2Former} & R-50 & ImageNet & 5.5 \\
& & & R-50 & COCO & 9.6 \\
& & & Swin-T & COCO & 9.8 \\
& & & Swin-L & COCO & 13.4 \\ \midrule
\multirow{5}*{ENT} & \multirow{5}*{AP$^e$} & Mask-RCNN & R-50 & ImageNet & 24.9 \\ 
& & \multirow{4}*{Mask2Former} & R-50 & ImageNet & 26.0 \\
& & & R-50 & COCO & 35.2 \\
& & & Swin-T & COCO & 42.9 \\
& & & Swin-L & COCO & 46.2 \\ \bottomrule

\end{tabular}
\end{table}

\vspace{1mm}
\noindent \textbf{Benchmark on EntitySeg Dataset.} In Table~\ref{Tab:aba_benchmark_merge}, we present results on semantic, instance, panoptic, and entity segmentation on the EntitySeg dataset with various baseline methods. 
Similar to existing segmentation datasets such as COCO and ADE20K, leveraging COCO pre-trained weights and a stronger backbone proves beneficial for enhancing each segmentation performance on the EntitySeg dataset. 
Compared to COCO and ADE20K, the EntitySeg dataset is more challenging as evidenced by the smaller performance numbers (\eg, Swin-L Mask2Former's PQ on COCO-PAN is 57.8\%).

\vspace{1mm}
\noindent \textbf{Transferring EntitySeg to Other General Segmentation Tasks.}
Table~\ref{Tab:aba_transfer_general_seg} shows the performance evaluation of transferring models pretrained on either COCO or EntitySeg dataset to other image/video-level segmentation tasks. 
Owing to its emphasis on in-the-wild generalization, EntitySeg provides greater improvements than COCO.
\begin{table}[t!]
\centering
\scriptsize
\caption{\textbf{Relative performance gains of EntitySeg pretraining over COCO-Panoptic pretraining when fine-tuned on various image/video-level segmentation datasets.}}
\label{Tab:aba_transfer_general_seg}
\setlength\tabcolsep{3pt}
\begin{tabular}{cccccc}
\cellcolor{lightgray!30} Transfer Dataset &
\cellcolor{lightgray!30} Type &
\cellcolor{lightgray!30} SEM (mIoU) &
\cellcolor{lightgray!30} INS (AP) &
\cellcolor{lightgray!30} PAN (PQ) & 
\cellcolor{lightgray!30} ENT (AP$^e$)\\ \toprule
ADE20K & Image & 57.2\textcolor[RGB]{34,139,34}{(+1.0)} & 36.0\textcolor[RGB]{34,139,34}{(+1.1)} & 
49.0\textcolor[RGB]{34,139,34}{(+0.9)} & 40.6\textcolor[RGB]{34,139,34}{(+1.5)} \\ 
YVIS2019~\cite{yang2019video} & Video & - & 61.3\textcolor[RGB]{34,139,34}{(+0.9)} & - & -\\
VIPSeg~\cite{miao2022large} & Video & - & - & 37.0\textcolor[RGB]{34,139,34}{(+1.1)} & - \\
UVO~\cite{wang2021unidentified} & Video & - & - & - & 29.5\textcolor[RGB]{34,139,34}{(+2.2)}\\ \bottomrule
\end{tabular}
\end{table}

\newcommand{\myarrow}[1][1cm]{\mathrel{
   \vcenter{\hbox{\rule[+.1pt]{#1}{.35pt}}}
   \mkern-10mu\hbox{\usefont{U}{lasy}{m}{n}$\longrightarrow$}}}

\begin{table*}[t!]
\centering
\scriptsize
\caption{\textbf{Comparison with high-resolution methods on EntitySeg with multi-scale training.} ``(x, y)'' indicates number of pixels along the shorter (x) and longer (y) sides of image. Rows 1, 2, \& 3 are single-scale Mask2Formers trained and evaluated at various single-scale resolutions. (V-Mask2Former \& VITA) mask predictions from full image and four crops are fused with either video-level Mask2Former or VITA.}
\label{Tab:aba_cropformer_cost}
\setlength{\tabcolsep}{3pt}
\begin{tabular}{clcccccc}
\cellcolor{lightgray!30} Method &
\cellcolor{lightgray!30} \Gape[0pt][2pt]{\makecell{Input Preprocessing}} &
\cellcolor{lightgray!30} GPU Type &
\cellcolor{lightgray!30} AP$^e$ &
\cellcolor{lightgray!30} \makecell[c]{GPU Memory \\ (Train)} &
\cellcolor{lightgray!30} \makecell[c]{GPU Memory \\(Inference)} &
\cellcolor{lightgray!30} \makecell[c]{Train Time \\(GPU day)} &
\cellcolor{lightgray!30} \makecell[c]{Inference Time \\(ms)} \\ \toprule
Mask2Former & high-res image $\stackrel{\text{resize}}{\longrightarrow}$ (800, 1333) & A100-40G & 39.5 & 5.8G & 3.0G & 2.4 & 637\\
Mask2Former & high-res image $\stackrel{\text{resize}}{\longrightarrow}$ (\textbf{2700}, \textbf{4500}) & A100-40G & OOM & OOM & - & OOM & - \\
Mask2Former & high-res image $\stackrel{\text{resize}}{\longrightarrow}$ (\textbf{2700}, \textbf{4500}) & \textbf{A100-80G} & 40.9 & \textbf{58.8G} & \textbf{10.4G} & \textbf{60.8} & \textbf{3312} \\ \midrule 
V-Mask2Former~\cite{cheng2021mask2former} & \multirow{3}{*}{high-res image \begin{tabular}{@{}c@{}} $\stackrel{\text{resize}}{\myarrow[59px]}$ (800, 1333) \\ \hspace{-2px}$\underset{\text{crop(}\delta\text{)}}{\longrightarrow}$ high-res crop $\underset{\text{resize}}{\longrightarrow}$ (800, 1333) \end{tabular}} & A100-40G & 39.7 & 10.4G & 6.7G & 6.8 & 1503\\
VITA~\cite{heo2022vita} & & A100-40G & 39.9 & 11.9G & 6.8G & 10.2 & 1512\\
CropFormer  & &  A100-40G &\textbf{41.0} & 11.4G & 6.9G & 9.8 & 1503 \\ \bottomrule
\end{tabular}
\end{table*}

\noindent \textbf{Benefit of EntitySeg for High-Quality Image Segmentation.}
In Table~\ref{Tab:aba_transfer_highquality_seg}, we transfer Mask2Former pretrained on EntitySeg dataset for finetuning on Dichotomous Image Segmentation~\cite{qin2022highly}. 
Compared to state-of-the-art ISNet~\cite{qin2022highly} that is meticulously designed for the task, Mask2Former pretrained on EntitySeg outperforms ISNet by a large margin, despite not having any task-specific design. 
These results demonstrate the merits of high-quality mask annotations in the EntitySeg dataset.
\begin{table}[t!]
\centering
\scriptsize
\caption{\textbf{Evaluation on Dichotomous Image Segmentation (DIS5K dataset~\cite{qin2022highly}).} ``Ours'' indicates Mask2Former pretrained on EntitySeg and then finetuned on DIS5K.}
\label{Tab:aba_transfer_highquality_seg}
\setlength\tabcolsep{3pt}
\begin{tabular}{ccccccc}
\cellcolor{lightgray!30} Method &
\cellcolor{lightgray!30} $F_{\beta}^{mx} \uparrow$ &
\cellcolor{lightgray!30} $F_{\beta}^{w} \uparrow$ & 
\cellcolor{lightgray!30} $M \downarrow$ &
\cellcolor{lightgray!30} $S_{\alpha} \uparrow$ &
\cellcolor{lightgray!30} $E_{\phi}^{m} \uparrow$ &
\cellcolor{lightgray!30} $HCE_{\gamma} \downarrow$
\\ \toprule
IS-Net~\cite{qin2022highly} & .791 & .717 & .074 & .813 & .856 & 1116 \\
Ours & \textbf{.872} & \textbf{.789} & \textbf{.035} & \textbf{.896} & \textbf{.913} & \textbf{969}\\ \bottomrule
\end{tabular}
\end{table}

\vspace{1mm}
\noindent \textbf{Segmentation In The Wild.}
Table~\ref{Tab:aba_transfer_wild_seg} compares in-the-wild segmentation performance between Mask2Formers trained on either EntitySeg or COCO dataset. 
Here, we directly use the trained models without finetuning on the evaluation datasets. 
The EntitySeg model outperforms the COCO-Panoptic \cite{kirillov2019panoptic} and COCO-Entity \cite{qi2021open} schemes by large margins.
%


\begin{table}[t!]
\centering
\scriptsize
\caption{\textbf{In-the-wild segmentation performance (AR@100 with IoU threshold 0.5).} We directly evaluate models trained on either EntitySeg or COCO on large-vocabulary segmentation LVIS~\cite{gupta2019lvis}, object clutter indoor segmentation for robot grab~\cite{DBLP:conf/icra/SuchiPFV19}, camouflaged instance segmentation CAMO~\cite{le2021camouflaged}, and few-shot segmentation FSS~\cite{li2020fss}. All models are based on Mask2Former with a Swin-Large backbone.}
\label{Tab:aba_transfer_wild_seg}
\setlength\tabcolsep{1pt}
\begin{tabular}{ccccccc}
\cellcolor{lightgray!30} Train Dataset & 
\cellcolor{lightgray!30} Task & 
\cellcolor{lightgray!30} LVIS~\cite{gupta2019lvis} &
\cellcolor{lightgray!30} OCID~\cite{DBLP:conf/icra/SuchiPFV19} &
\cellcolor{lightgray!30} CAMO~\cite{le2021camouflaged} &
\cellcolor{lightgray!30} FSS~\cite{li2020fss} &
\cellcolor{lightgray!30} Avg. AR\\ \toprule
COCO & PAN & 24.38 & 39.94 & 43.80 & 64.96 & 43.27 \\
COCO & ENT & 27.75 & 45.88 & 59.70 & 75.60 & 52.23 \\ \midrule
SAM-H~\cite{kirillov2023segment} & Prompt & \textbf{49.25} & 55.17 & \textbf{75.33} & 73.60 & 63.34 \\ 
EntitySeg & ENT & \textbf{33.56} & \textbf{76.63} & 73.20 & \textbf{86.26} & \textbf{67.41} \\ \bottomrule
\end{tabular}
\end{table}

\subsection{Comprehensive Analysis of CropFormer}
We first ablate the Cropformer model on the entity segmentation task and then report results on other datasets.

Table~\ref{Tab:aba_cropformer_cost} shows the benefits of CropFormer on segmentation performance and computational cost compared to the Mask2Former~\cite{cheng2022masked}, its multi-scale extension, as well as related methods
~\cite{cheng2021mask2former,heo2022vita} that use multi-view fusion. 
The first four rows are our baseline Mask2Former with the single-scale inference (800, 1040, or 2700 for the shorter side) and (1333, 1732 or 4500 for the longer side) for the full images, where 1040 = 800 $\times \delta$ and 1732 = 1333 $\times \delta$ with $\delta$ = 0.7 by default. 
For the results in the fifth row, we use test-time Hungarian matching to associate the same entities obtained with multi-scale images.
This inference strategy improves our results minimally. 
In the sixth and seventh rows, directly using video-level Mask2Former or VITA, the state-of-the-art video instance segmentation framework, merely brings marginal performance gains.
In contrast, using the crop output from the batch decoder of CropFormer achieves a significant AP$^\text{e}$ gain as indicated by the last row. 
Combining the full image and the four crop outputs from CropFormer (last row) yields even more substantial 1.5 AP$^\text{e}$ gains compared to the baseline (first row). 
In particular, compared to Mask2Former baselines with a straightforward increase in image resolution, CropFormer demonstrates significantly enhanced computational efficiency.

For more ablation studies on CropFormer such as training schedules, decoder type and crop dataloader, please refer to our appendix.


\subsection{GELLA on Entity Grounding}
\begin{table*}[!t]
\centering
\footnotesize
\vspace{-2mm}
\caption{\textbf{Referring Expression Segmentation Results.} Our performance across RefCOCO, RefCOCO+, and RefCOCOg in generating accurate segmentation masks based on referring expressions surpasses that of closely related work. $\star$ indicates that the method used the GranD dataset [85] for pretraining, which is significantly larger than
the datasets used by other methods}
\vspace{-2mm}
\label{tab:results_refer}
\begin{tabular}{lccccccccccc}
\toprule
\cellcolor{lightgray!30} 
& 
\cellcolor{lightgray!30}  
& 
\multicolumn{3}{c}{\cellcolor{lightgray!30} RefCOCO} && 
\multicolumn{3}{c}{\cellcolor{lightgray!30} RefCOCO+} 
&& \multicolumn{2}{c}{\cellcolor{lightgray!30} RefCOCOg} \\
\cline{3-5} \cline{7-9} \cline{11-12}
\multirow{-2}{*}{\cellcolor{lightgray!30} \makecell[c]{Method}} &
\multirow{-2}{*}{\cellcolor{lightgray!30} Venue} & 
\cellcolor{lightgray!30} val & 
\cellcolor{lightgray!30} testA & 
\cellcolor{lightgray!30} testB && 
\cellcolor{lightgray!30} val & 
\cellcolor{lightgray!30} testA & 
\cellcolor{lightgray!30} testB && 
\cellcolor{lightgray!30} val(U) & 
\cellcolor{lightgray!30} test(U) \\
\midrule
CRIS~\cite{wang2022cris} & CVPR2022 & 70.5 & 73.2 & 66.1 && 65.3 & 68.1 & 53.7 && 59.9 & 60.4 \\
LAVT~\cite{yang2022lavt} & CVPR2022 & 72.7 & 75.8 & 68.8 && 62.1 & 68.4 & 55.1 && 61.2 & 62.1 \\
GRES~\cite{liu2023gres} & CVPR2023 & 73.8 & 76.5 & 70.2 && 66.0 & 71.0 & 57.7 && 65.0 & 66.0 \\
X-Decoder~\cite{zou2023generalized} & CVPR2023 & - & - & - && - & - & - && 64.6 & - \\
SEEM~\cite{kirillov2023segment} & NeurIPS2023 & - & - & - && - & - & - && 65.7 & - \\ \midrule
LISA-7B~\cite{lai2023lisa} & CVPR2024 & 74.9 & 79.1 & 72.3 && 65.1 & 70.8 & 58.1 && 67.9 & 70.6 \\
GLaMM$^\star$-13B~\cite{hanoona2023GLaMM} & CVPR2024 & \textbf{79.5} & - & - && 66.3 & - & - && 69.1 & - \\
PixelLM~\cite{ren2023pixellm} & CVPR2024 & 73.0 & 76.5 & 68.2 && 66.3 & 71.7 & 58.3 && 69.3 & 70.5\\
\midrule
GELLA-7B & - & 76.1 & 79.9 & 73.1 && 66.4 & 72.7 & 60.2 && 69.6 & 71.3 \\
GELLA-13B & - & 76.7 & \textbf{80.5} & \textbf{73.6} && \textbf{67.0} & \textbf{73.2} & \textbf{60.6} && \textbf{70.4} & \textbf{71.5} \\
\bottomrule
\end{tabular}
\end{table*}

We quantitatively evaluate GELLA on three tasks: panoptic segmentation, referring expression segmentation, and panoptic narrative grounding.
We first present experimental results for referring expression segmentation and panoptic narrative grounding to show the effectiveness of our method. 
Then, we mainly ablate our colormap design without a language branch on panoptic segmentation. 
Finally, we provide qualitative visualization results of our GELLA framework.

We initialize the language branch of our framework with the fully fine-tuned LLaVA-v1.5~\cite{liu2023improvedllava} pre-trained model, coupled with a CLIP~\cite{radford2021learning} vision encoder. 
The CLIP component specifically utilizes the ViT-Large-336 architecture as its backbone. 
Additionally, we adopt MobileNetV2~\cite{sandler2019mobilenetv2} as the backbone of our mask encoder, following the design of image encoder in Mask2Former.


We choose the Panoptic Narrative Grounding (PNG) dataset~\cite{gonzalez2021panoptic} as the foundational data for our experiments. 
To ensure fair comparison across the three tasks, we train our model using images from the respective training sets of each task, totaling 10,813 images. 
In addition to the PNG dataset for training the segmentation branch, we use GPT-4~\cite{openai2023gpt4} to generate two descriptive captions for each image in the EntitySeg training dataset~\cite{qi2023high}, which contains approximately 31,000 images. 
Unlike the COCO dataset~\cite{lin2014microsoft}, EntitySeg encompasses various image domains, including indoor and outdoor environments, street scenes, cartoons, and aerial imagery.

A crucial data augmentation protocol applied across all experiments involves resizing input images to the target output dimensions. 
The same shorter and longer sizes constrain the output dimensions, and we pad the resized image to a square with zero values. 
This method aligns with the image dimensions used during the pretraining phase of the CLIP vision encoder, typically 224 or 336 pixels. 
To preserve entity integrity, we opt against employing cropping techniques that risk truncating entities.

We train our method on 8 A100 GPUs for 50 epochs with an initial learning rate of $10^{-5}$ and the AdamW optimizer with standard parameters. 
The learning rate decayed by 0.1 after 46 and 48 epochs, respectively. 
During each training iteration, the sample ratios for image description, semantic noun extraction, and narrative grounding are 0.2, 0.2, and 0.6. The batch size is 16. 
The constraints for the shorter and longer sizes of the CLIP vision encoder are 336. 
The colormaps used in the inference stage are from the class-agnostic entity segmentation results of CropFormer with a Swin-Large backbone, the state-of-the-art entity segmentation method. We note that the main experiments and ablation studies of the ESG pipeline are conducted on this setting.

\begin{table}
\centering
\footnotesize
\vspace{-2mm}
\caption{\textbf{Panoptic Narrative Grounding Results.} Performance on PNG dataset in generating segmentation masks based on a long caption.}
\vspace{-2mm}
\label{tab:results_png}
\setlength\tabcolsep{3pt}
\begin{tabular}{lcccccc}
\toprule
\cellcolor{lightgray!30} Method & 
\cellcolor{lightgray!30} Venue & \cellcolor{lightgray!30} AR & \cellcolor{lightgray!30} AR$_{\text{Th}}$ & \cellcolor{lightgray!30} AR$_{\text{St}}$ & \cellcolor{lightgray!30} AR$_{\text{Sing}}$ & \cellcolor{lightgray!30} AR$_{\text{Pl}}$ \\ \midrule
MCN~\cite{margffoy2018dynamic} & CVPR2020 & - & 48.2 & - & - & - \\
PNGb~\cite{gonzalez2021panoptic} & ICCV2021 & 55.4 & 56.2 & 54.3 & 56.2 & 48.4 \\
PPMN~\cite{ding2022ppmn} & MM2022 & 59.4 & 57.2 & 62.5 & 60.0 & 50.4 \\
NICE~\cite{wang2023nice} & MM2023 & 62.3 & 60.2 & 65.3 & 63.1 & 55.2\\
PiGLET~\cite{gonzalez2023piglet} & TPAMI2023 & 65.9 & 64.0 & 68.6 & 67.2 & 54.5 \\
PPO-TD~\cite{hui2023enriching} & IJCAI2023 & 66.1 & 64.0 & 70.7 & 68.1 & 58.3 \\
GSVA~\cite{xia2024gsva} & CVPR2024 & 68.5 & 64.7 & 71.1 & 68.5 & 58.4 \\
OMG-LLAVA~\cite{OMGLLaVA} & NeurlPS2024 & 68.7 & 65.1 & 71.3 & 68.9 & 58.8\\
\midrule
GELLA-7B & - & 69.8 & 66.2 & 72.3 & 69.9 & 58.4\\
GELLA-13B & - & \textbf{71.3} & \textbf{67.8} & \textbf{73.1} & \textbf{71.0} & \textbf{59.5} \\
\bottomrule
\end{tabular}
\end{table}

\subsubsection{Referring Expression Segmentation}
We evaluate our model on the RefCOCO~\cite{yu2016modeling}, RefCOCO+~\cite{yu2016modeling}, and RefCOCOg~\cite{mao2016generation,nagaraja2016modeling} benchmarks for referring expression segmentation. 
The prompt is structured as follows: ``Please help me extract each semantic noun from the sentence. The sentence is $<$EXPRESSION$>$'', with $<$EXPRESSION$>$ filled by the corresponding text from the validation set. 
As shown in Table~\ref{tab:results_refer}, our model demonstrates considerable performance in all datasets despite not being explicitly trained on the captions of the datasets. 
This suggests that the underlying large language model has a robust ability to comprehend text. 
In the referring expression segmentation task, where each sentence typically references a single main entity, our model can effectively ignore additional attributive nouns because our association module assigns lower scores to irrelevant segmentation masks. 
Scaling up the language model from 7B to 13B parameters does not improve grounding accuracy. 
This implies that the bottleneck may lie in the segmentation branch rather than language understanding.

Furthermore, we compare our model's performance with GLAMM-13B~\cite{hanoona2023GLaMM} and PixelLM~\cite{ren2023pixellm} on RefCOCO, RefCOCO+, and RefCOCOg. 
The results show that our model outperforms GLAMM-13B and PixelLM on RefCOCO+ and RefCOCOg but falls short on RefCOCO. 
This discrepancy could be attributed to differences in dataset annotations. 
Captions in RefCOCO+ and RefCOCOg are generally longer and emphasize attributes and implicit spatial localization, unlike the shorter, more direct captions in RefCOCO. 
Consequently, the superior performance on RefCOCO+ and RefCOCOg highlights the reasoning capabilities of our GELLA model. 
Similar observations are discussed in PixelLM~\cite{ren2023pixellm}.

\subsubsection{Panoptic Narrative Grounding}
Table~\ref{tab:results_png} presents the quantitative performance of the GELLA framework on panoptic narrative grounding (PNG).
This task surpasses the complexity of referring expression segmentation, requiring the segmentation of every semantic noun from extensive captions, demanding a comprehensive grasp of the entire narrative.
The deployment of a large language model (LLM) is pivotal in this scenario. Using the same prompts as in referring expression segmentation offers substantial improvements in task execution. 
GELLA achieves a significant gain with a 7B parameter language branch, enhancing the Average Recall (AR) by 3.7 points beyond the leading method, PPO-TD. 
Progressing to an even larger language model, such as one with 13B parameters, consistently enhances performance. 
These results underscore the importance of LLM capacity for parsing lengthy paragraphs, a skill that evaluating referring expression segmentation with brief sentences fails to fully leverage.
%
In addition, the proposed pipeline outperforms both GSVA and OMG-LLAVA. 
This improvement is attributed to the high-quality entity segmentation results achieved by our decoupled two-stage pipeline, which provides higher recall for the grounding performed by the GELLA model.

In Table~\ref{rebutal_computation_cost_1} and~\ref{rebutal_computation_cost_2}, we compare our method to the concurrent works of LISA and GLaMM on COCO panoptic narrative grounding. 
Our method outperforms other alternatives with a lower computational cost, even when accounting for the cost of segmentation masks generated by Mask2Former with a Swin-Large backbone (additional 218M hyper-parameters, 541GFLOPS, and 0.3s inference time).

Moreover, GLAMM builds an automated annotation pipeline to label the SA-1B dataset using SAM and GPT-4 whereas our ESG pipeline is based on human-labeled data. From the perspective of the common training paradigm of pretraining followed by supervised fine-tuning (SFT), we attribute our improvement to role of our proposed data. Our EntitySeg dataset provides fine-grained mask annotations for all entities present in each image, whereas images in SA-1B are typically annotated with only a small subset of entities, making them more suitable for pretraining than for SFT.

\begin{table}[t!]
\caption{\textbf{Comparison of detailed structures and corresponding hyper-parameter numbers indicated in $(\_\_)$.} For fair comparisons, we use the LLAMA-7B as our language decoder. `SAM' and `CLIP' indicate the adopted ViT-Huge and ViT-Large backbone. For the decoder part, `X-Style' means the corresponding decoder design of X.}
\label{rebutal_computation_cost_1}
\centering
\scriptsize
\setlength{\tabcolsep}{1pt}
\begin{tabular}{c|c|c|c|c}
\toprule
\cellcolor{lightgray!30} Method &
\cellcolor{lightgray!30} Image Encoder & 
\cellcolor{lightgray!30} Mask Encoder & 
\cellcolor{lightgray!30} Decoder \\ \midrule
LISA & SAM (\underline{641M}) + CLIP (\underline{300M}) & - & SAM-Style (4M) \\
GLaMM & SAM (\underline{641M}) + CLIP (\underline{300M}) & - & Mask2Former-Style (21M) \\
Ours &  CLIP-ViT-L (\underline{300M}) & MobileNetV2 (\underline{3.5M}) & Mask2Former-Style (21M) \\
\bottomrule
\end{tabular}
\end{table}

\begin{table}[t!]
\caption{\textbf{Comparison of GFLOPS and average inference seconds per image.}`LLM', `IME', and `MMD' denote the large language model, the together of image and mask encoder, and the mask decoder in GELLA, respectively. `ALL' denotes the full GELLA model, which includes `LLM', `IME', and `MMD'. `AR' is short for average recall, which indicates inference performance.
}
\label{rebutal_computation_cost_2}
\begin{minipage}{\textwidth}
        \begin{minipage}[t]{0.22\textwidth}
        \centering
        \scriptsize
        \setlength{\tabcolsep}{2pt}
        \begin{tabular}{ccccc}
        \toprule
        \cellcolor{lightgray!30} Method &
        \cellcolor{lightgray!30} LLM &
        \cellcolor{lightgray!30} IME & 
        \cellcolor{lightgray!30} MMD &
        \cellcolor{lightgray!30} ALL
        \\ \midrule
        LISA & 92607 & 3114 & 226 & 95947 \\
        GLaMM & 11987 & 3114 & 113 & 15214 \\
        Ours & 11987 & 204 & 113 & 12304 \\
        \bottomrule
        \end{tabular}
        
    (a) GFLOPS.
        \end{minipage}
        \begin{minipage}[t]{0.31\textwidth}
        \centering
        \scriptsize
        \setlength{\tabcolsep}{1pt}
        \begin{tabular}{cccccc}
        \toprule
        \cellcolor{lightgray!30} Method &
        \cellcolor{lightgray!30} LLM & 
        \cellcolor{lightgray!30} IME & 
        \cellcolor{lightgray!30} MMD &
        \cellcolor{lightgray!30} ALL &
        \cellcolor{lightgray!30} AR
        \\ \midrule
        LISA & \multirow{3}*{3.2} & 1.9 & 0.2 & 5.5 & 43.6\\
        GLaMM & & 1.9 & 0.3 & 5.4 & 63.1 \\
        Ours & & 0.7 & 0.3 & 4.2 & 69.8\\
        \bottomrule
        \end{tabular}
        
     (b) Inference time.
    \end{minipage}

\end{minipage}
\end{table}

\subsubsection{Panoptic Segmentation}
Although the GELLA framework is designed for narrative grounding tasks, its architecture allows for generalization to conventional panoptic segmentation. 
During inference, we employ a comprehensive prompt that enumerates all 133 category names, structured as: ``There might exist a person, dog, cat, ... and window.'' Similar to referring expression segmentation, entities within the image corresponding to the listed category nouns are identified. 
In contrast, categories not present in the image are assigned a deficient score, effectively filtering out entities from the segmentation output. 

\begin{table}[t!]
\centering
\footnotesize
\vspace{-2mm}
\caption{\textbf{Closed and Open Panoptic Segmentation Results.} Performance on COCO and ADE20K panoptic segmentation dataset in generating segmentation masks based on all categories list. For the ADE20K, it is the cross-dataset evaluation due to the model is trained in COCO dataset.}
\vspace{-2mm}
\label{tab:results_pan}
\setlength\tabcolsep{3pt}
\begin{tabular}{ccccccc}
\toprule
\cellcolor{lightgray!30} Method & 
\cellcolor{lightgray!30} Venue & 
\cellcolor{lightgray!30} Epoch & 
\cellcolor{lightgray!30} PQ & 
\cellcolor{lightgray!30} SQ & 
\cellcolor{lightgray!30} RQ & 
\cellcolor{lightgray!30} PQ (ADE20K) \\ \midrule
\multirow{2}{*}{Mask2Former~\cite{cheng2022masked}} & \multirow{2}{*}{CVPR2022} & 50 & 53.4 & 83.1 & 63.4 & - \\
& & 12 & 49.2 & 82.5 & 58.8 & - \\ 
MaskCLIP~\cite{ding2022open} & ICML2023 & 50 & - & - & - & 15.1 \\
OPSNet~\cite{chen2023open} & ICCV2023 & 50 & - & - & - & 17.7 \\
\midrule
\multirow{2}{*}{GELLA-7B} & \multirow{4}{*}{-} & 12 & 53.1 & 83.3 & 63.8 & 18.1 \\
& & 50 & 56.4& 83.4 & 66.8 & 20.5 \\
\multirow{2}{*}{GELLA-13B} & & 12 & 53.2 & 83.2 & 63.9 & 19.8 \\
& & 50 & \textbf{56.5} & \textbf{83.3} & \textbf{67.3} & 21.1 \\ \bottomrule
\end{tabular}
\end{table}

Table~\ref{tab:results_pan} shows the performance of our GELLA framework on the panoptic segmentation task within the COCO dataset. 
Our framework, which deviates from the conventional approach by not using a traditional image encoder with a Swin-Tiny backbone, achieves improved results. 
However, we note that this comparison may not be entirely equitable since our framework's colormaps are derived from a Swin-Large backbone. 
Despite this, we posit that the results still validate the sufficiency of low-resolution images for the CLIP vision encoder regarding classification accuracy. 
Furthermore, scaling up the language branch does not enhance segmentation performance. 
This suggests that extracting semantic nouns is not a bottleneck in this context, and additional capacity in the language branch does not contribute to significant gains.

Furthermore, some open-vocabulary segmentation methods~\cite{ding2022open,chen2023open} that crop the mask region and then pool features fail to preserve the context features and positional information across the entire feature map. This enables more accurate and general association for segmentation in the wild, especially when multiple entities belong to the same category.

\begin{table}[t!]
\caption{
\textbf{Comparison on reasoning and zero-shot segmentation.} (a) The performance comparison of reasoning segmentation on ReasonSeg evaluation dataset proposed by LISA~\cite{lai2023lisa}. 
(b) The performance comparison of open-vocabulary segmentation on SGinW~\cite{ren2024grounded} benchmark within zero-shot setting.}
\label{tab:abl_reason_open}
\begin{minipage}{\textwidth}
\begin{minipage}[t]{0.25\textwidth}
        \centering
        \footnotesize
        \setlength{\tabcolsep}{2pt}
        \begin{tabular}{ccccc}
        \toprule
            \cellcolor{lightgray!30} 
            & \multicolumn{2}{c}{\cellcolor{lightgray!30} val} & \multicolumn{2}{c}{\cellcolor{lightgray!30} test} \\ \cline{2-5} 
            \multirow{-2}{*}{\cellcolor{lightgray!30} 
            Method} & \cellcolor{lightgray!30} gIoU & 
            \cellcolor{lightgray!30} cIoU & \cellcolor{lightgray!30} gIoU & \cellcolor{lightgray!30} cIoU
            \\ \midrule
            LISA~\cite{lai2023lisa} & 65.0 & 72.9 & 61.3 & 62.2 \\
            GLAMM~\cite{hanoona2023GLaMM} & 65.5 & 73.2 & 61.5 & 62.4 \\
            GELLA & \textbf{66.7} & \textbf{74.3} & \textbf{63.3} & \textbf{64.1}
            \\ \bottomrule
        \end{tabular}
        
    (a)Reasoning segmentation
        \end{minipage}
        \begin{minipage}[t]{0.26\textwidth}
        \centering
        \footnotesize
        \setlength{\tabcolsep}{2pt}
        \begin{tabular}{cccc}
        \toprule
        \cellcolor{lightgray!30} Method & \cellcolor{lightgray!30} AP \\ \midrule
        Grounded-SAM~\cite{ren2024grounded} & 48.7 \\
        Grounded-HQ-SAM~\cite{ke2024segment} & 49.6\\
        GELLA & \textbf{50.7} \\ \bottomrule 
        \end{tabular}
        
    (b)Open-vocabulary segmentation
    \end{minipage}
\end{minipage}
\end{table}

\subsubsection{Reasoning Segmentation}
In Table~\ref{tab:abl_reason_open}(a), we present performance comparisons with LISA and GLAMM on the reasoning segmentation task. 
%
For fair comparisons, we incorporate the fine-tuned data used in LISA to train GLAMM and our GELLA. 
The results of using the 13B large language model demonstrate that our method can easily fit the reasoning segmentation task, even without additional class-aware segmentation datasets like ADE20K, which is used in LISA. 
This indicates that the association module in the GELLA model effectively reduces the dependence on large-scale semantic-aware data.

\subsubsection{Zero-Shot Segmentation}
In Table~\ref{tab:abl_reason_open}(b), we compare the series of works~\cite{ren2024grounded,ke2024segment} derived from Grounded-SAM on the segmentation in the Wild (SGinW) zero-shot benchmark, which includes 25 in-the-wild zero-shot datasets. 
%
To account for the extensive training dataset used by Grounded-SAM, we augment the training process with most of the large-scale public segmentation datasets, including COCO~\cite{lin2014microsoft}, ADE20K~\cite{zhou2017scene}, Lvis~\cite{gupta2019lvis}, SA-1B~\cite{kirillov2023segment}, and AS-1B~\cite{wang2023all} (a joint set of AS-V2 and AS-100M). 
The results Table~\ref{tab:abl_reason_open} demonstrate that our model outperforms both Grounded-SAM~\cite{ren2024grounded} and Grounded-HQ-SAM~\cite{ke2024segment} when using large-scale datasets. 
However, we acknowledge that these comparisons may not be entirely fair as aligning the exact training data sets used is extremely challenging.

\subsubsection{Ablation Studies}
\label{sec:ablation}
Considering the experimental results across three tasks, we analyze the specific effect of the ResoBlend module, the colormap design and a CLIP vision encoder on panoptic segmentation.
Those ablation study allows us to better understand the effect of such three kinds of design on mask quality. 
Furthermore, we conduct experiments on panoptic narrative grounding to better understand the influence of the sampling strategy and the association module.

\begin{table}[t!]
\caption{
\textbf{Comparison of (a) image resolution used in CLIP vision encoder (b) Color assignment criterion in generating colormaps.}}
\label{tab:abl_clip_colormap}
\begin{minipage}{\textwidth}
\begin{minipage}[t]{0.22\textwidth}
        \centering
        \footnotesize
        \setlength{\tabcolsep}{2pt}
        \begin{tabular}{cccc}
        \toprule
            \cellcolor{lightgray!30} Resolution & 
            \cellcolor{lightgray!30} PQ & 
            \cellcolor{lightgray!30} SQ & 
            \cellcolor{lightgray!30} RQ
            \\ \midrule
            224 & 51.5 & 82.6 & 61.6\\
            336 & 53.1 & 83.3 & 63.8 
            \\ \bottomrule
        \end{tabular}
        
    (a)
        \end{minipage}
        \begin{minipage}[t]{0.23\textwidth}
        \centering
        \footnotesize
        \setlength{\tabcolsep}{2pt}
        \begin{tabular}{cccc}
        \toprule
        \cellcolor{lightgray!30} Color & 
        \cellcolor{lightgray!30} PQ & 
        \cellcolor{lightgray!30} SQ & 
        \cellcolor{lightgray!30} RQ \\ \midrule
        Location & 52.9 & 83.4 & 63.6 \\
        Random & 53.1 & 83.3 & 63.8 \\ \bottomrule 
        \end{tabular}
        
    (b)
    \end{minipage}
\end{minipage}
\end{table}

\begin{table}[t!]
\vspace{-3mm}
\caption{\textbf{Ablation study of (a) various backbones in colormap encoder (b) different segmentation masks in training.} We evaluate mask quality in AP due to the class-agnostic property.}
\vspace{-3mm}
\label{tab:abl_structure_colormap}
\begin{minipage}{\textwidth}
\begin{minipage}[t]{0.24\textwidth}
        \centering
        \footnotesize
        \setlength{\tabcolsep}{2pt}
        \begin{tabular}{cccc}
        \toprule
            \cellcolor{lightgray!30} Backbone & 
            \cellcolor{lightgray!30} AP & 
            \cellcolor{lightgray!30} AP$_\text{50}$ & \cellcolor{lightgray!30} AP$_\text{75}$
            \\ \midrule
            MobileNet(V2) & 95.2 & 96.9 & 96.1\\
            Swin-Tiny & 95.8 & 98.9 & 97.3 \\
            Swin-Large & 96.3 & 99.0 & 97.8 \\
            \bottomrule
        \end{tabular}
        
    (a)
        \end{minipage}
        \begin{minipage}[t]{0.24\textwidth}
        \centering
        \footnotesize
        \setlength{\tabcolsep}{2pt}
        \begin{tabular}{cccc}
        \toprule
        \cellcolor{lightgray!30} Masks & 
        \cellcolor{lightgray!30} AP & 
        \cellcolor{lightgray!30} AP$_\text{50}$ & \cellcolor{lightgray!30} AP$_\text{75}$
        \\ \midrule
        COCO$^{\text{p}}$ & 95.2 & 98.7 & 97.1 \\
        Entity$^{\text{p}}$ & 94.9 & 98.5 & 97.0 \\
        COCO$^{\text{g}}$ & 95.8 & 98.9 & 97.3 \\ \bottomrule
        \end{tabular}
        
    (b)
    \end{minipage}
\end{minipage}
\end{table}

\vspace{1mm}
\noindent \textbf{CLIP Vision Encoder.}
Table~\ref{tab:abl_clip_colormap}(a) presents the results of our ablation study on the input resolution for the ViT backbone. 
Using a ViT model trained on lower image resolutions, such as 224 pixels, leads to performance loss. 
This decrease in accuracy is attributed to the model's capacity to effectively recognize small objects, which are prevalent in the COCO dataset.

\vspace{1mm}
\noindent \textbf{Colormap Design.}
Table~\ref{tab:abl_clip_colormap}(b) details the ablation study of various color assignment strategies for encoding entity segmentation masks. 
The first row explores color assignment based on the entity's gravity centroid, as described in Painter~\cite{wang2023images}. 
There is no significant performance difference between this centroid-based assignment and a random color assignment approach. 
This can be attributed to that the model can group pixels sharing the same color, regardless of the specific color used. 
Consequently, we opt for random color assignment throughout our experiments due to its simplicity and effectiveness.

Table~\ref{tab:abl_structure_colormap} shows the effectiveness of solely utilizing colormap encoding and decoding across various backbones and colormap sources. 
These sources encompass the segmentation results generated from models trained on the COCO or EntitySeg datasets and the ground-truth mask annotations from COCO. 
These results show that the colormap design maintains its robustness irrespective of the backbone architecture and variability in mask patterns. 
The proposed approach to encoding and decoding colormaps is adaptive to different underlying structures and sources of segmentation information.

\vspace{1mm}
\noindent \textbf{ResoBlend Module.}
Table~\ref{tab:module_ResoBlend} shows the results of our ablation study on the various components within the proposed ResoBlend module. 
Our baseline configuration, detailed in the last row, achieves optimal performance. 
Omitting any of the modules degrades mask prediction quality. 
Notably, segmentation accuracy significantly drops when mask features are isolated from image features, as shown in the second row. This effect underscores the importance of feature interaction between the image and mask features.
Nevertheless, even with this separation, the network retains a limited capacity for entity classification. This is attributed to the language embeddings, which contain visual information linked to the CLIP feature and function as image tokens within a sentence. 
This design aspect enhances the language branch's ability to discern semantic nouns in the image context.

\begin{figure*}[t!]
\vspace{-2mm}
\caption{\textbf{Illustration of our user interface.} The GELLA framework can perform three tasks: image description, semantic noun extraction, and narrative grounding. We provide more visualization results in our appendix.}
\vspace{-2mm}
\label{fig.user_interface}
\centering
\includegraphics[width=\textwidth]{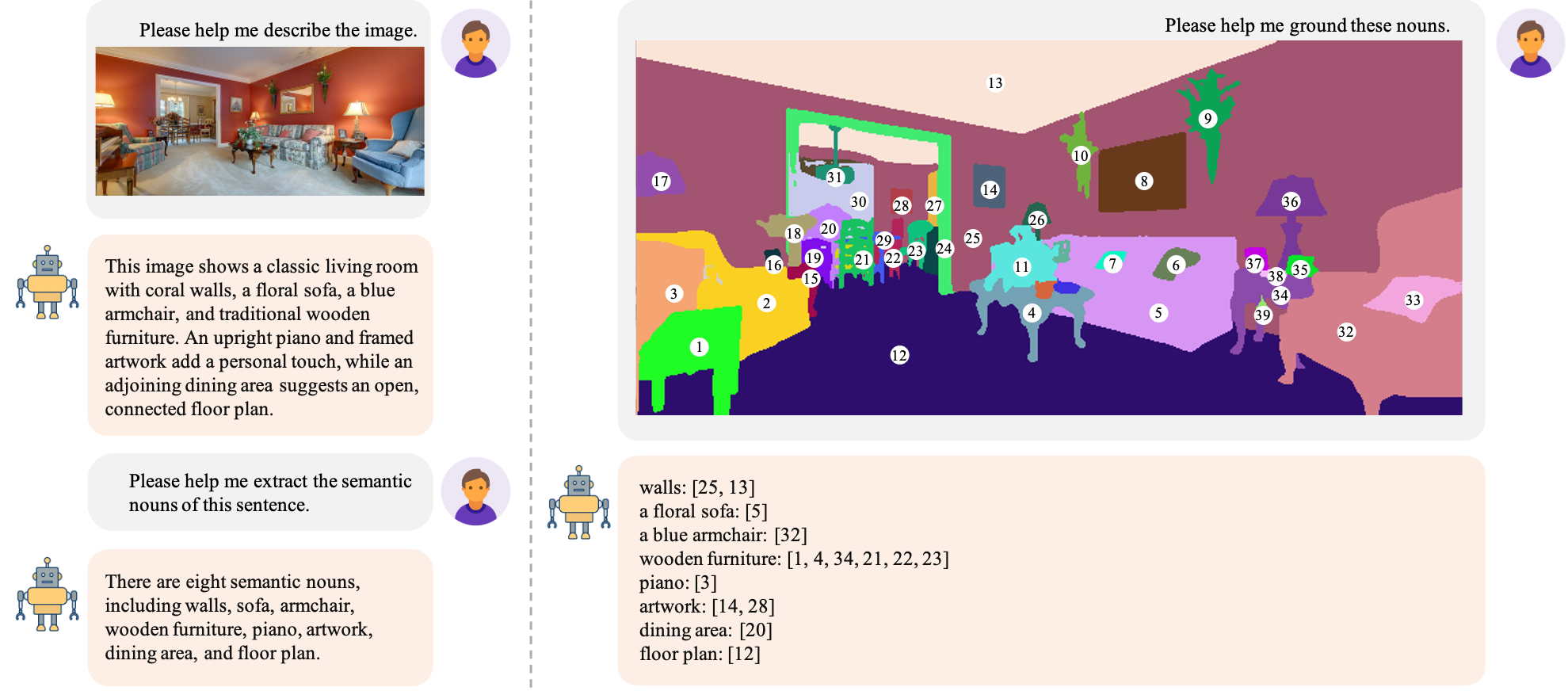}
\end{figure*}

\begin{table}[t!]
\centering
\footnotesize
\vspace{-2mm}
\caption{\textbf{Ablation study of modules in the proposed ResoBlend module.} ``SA'', ``CA'', ``FF'', ``RES'' and ``CONV'' abbreviate the self-attention, cross-attention, feed-forward, residual connection, and convolution module, as illustrated in the right part of Figure~\ref{fig.framework}.}
\vspace{-2mm}
\label{tab:module_ResoBlend}
\begin{tabular}{cccccccc}
\toprule
\cellcolor{lightgray!30} SA & 
\cellcolor{lightgray!30} CA & 
\cellcolor{lightgray!30} FF & 
\cellcolor{lightgray!30} RES & 
\cellcolor{lightgray!30} CONV & 
\cellcolor{lightgray!30} PQ & 
\cellcolor{lightgray!30} SQ & 
\cellcolor{lightgray!30} RQ \\ \midrule
& $\checkmark$ & $\checkmark$& $\checkmark$& $\checkmark$ & 51.8 & 82.7 & 61.3 \\
$\checkmark$ & & $\checkmark$& $\checkmark$& $\checkmark$ & 43.1 & 78.1 & 58.1 \\
$\checkmark$ & $\checkmark$ & & $\checkmark$ & $\checkmark$ & 52.8 & 83.0 & 63.3 \\
$\checkmark$ & $\checkmark$ & $\checkmark$& & $\checkmark$ & 50.9 & 81.9 & 60.5 \\
$\checkmark$ & $\checkmark$ & $\checkmark$& $\checkmark$& & 52.4 & 82.5 & 61.8\\
$\checkmark$ & $\checkmark$ & $\checkmark$& $\checkmark$& $\checkmark$ & 53.1 & 83.3 & 63.8 \\ \bottomrule
\end{tabular}
\end{table}

\begin{table}[t!]
\vspace{-2mm}
\caption{\textbf{Comparison of different segmentation masks used in (a) training and (b) testing.}}
\vspace{-2mm}
\label{tab:abl_train_test_entity}
\begin{minipage}{\textwidth}
\begin{minipage}[t]{0.22\textwidth}
        \centering
        \footnotesize
        \setlength{\tabcolsep}{2pt}
        \begin{tabular}{cccc}
        \toprule
            \cellcolor{lightgray!30} Train & 
            \cellcolor{lightgray!30} PQ & 
            \cellcolor{lightgray!30} SQ & 
            \cellcolor{lightgray!30} RQ
            \\ \midrule
            COCO & 53.0 & 83.1 & 63.8\\
            Entity & 52.9 & 83.2 & 63.7\\
            Ground truth & 53.1 & 83.3 & 63.8 
            \\ \bottomrule
        \end{tabular}
        
    (a)
        \end{minipage}
        \begin{minipage}[t]{0.24\textwidth}
        \centering
        \footnotesize
        \setlength{\tabcolsep}{2pt}
        \begin{tabular}{cccc}
        \toprule
        \cellcolor{lightgray!30} Test & 
        \cellcolor{lightgray!30} PQ & 
        \cellcolor{lightgray!30} SQ & 
        \cellcolor{lightgray!30} RQ \\ \midrule
        COCO & 53.1 & 83.3 & 63.8 \\
        Entity & 49.2 & 83.7 & 58.6\\
        Ground truth & 79.6 & 97.8 & 91.4 \\ \bottomrule
        \end{tabular}
        
    (b)
    \end{minipage}
\end{minipage}
\vspace{-2mm}
\end{table}

Table~\ref{tab:abl_train_test_entity} ablates the effect of using segmentation masks from various sources during the training and testing phases. 
Table~\ref{tab:abl_train_test_entity}(a) demonstrates that the origin of mask data in training does not significantly alter outcomes. 
These results can be attributed to our segmentation branch's loss design, which employs the input as the training target. 
Consequently, the proposed model learns an identity function without refining the mask quality. 
Table~\ref{tab:abl_train_test_entity}(b) illustrates the classification potential of our CLIP vision encoder when provided with ground truth masks during testing, serving as an indicator of the upper-bound performance limit. 
However, when applying segmentation predictions from a model trained on the EntitySeg dataset, a noticeable domain gap persists concerning the COCO dataset in the case of annotations.

\noindent \textbf{Other Mask Encoder and Decoder Alternatives.}
We conduct ablation studies on several mask encoder and decoder alternatives. At first, we directly explore encoding binary entity masks by MobileNetV2. And then we use global pooling to encode them into entity embeddings and apply cross-attention between the image features and the entity embeddings, followed by self-attention among the entity embeddings.

In Table~\ref{tab:mask_structure}, we ablate the mask value range of $\mathbf{M}^{r}$ (i.e., $\{0,1\}$, $\{-1,1\}$, or $\{0,255\}$), the pooling strategy (average pooling or a learnable convolutional layer or `CLS' token), and the number of attention blocks on the panoptic narrative grounding task, which involves grounding multiple entities, and find that all variants perform worse than our proposed method.

We further analyze this result by comparing our design with such an alternative. First, our GELLA in the ESG pipeline takes as input a single colormap transformed from multiple binary entity masks, enabling the backbone to capture all entities jointly before attention is applied. In contrast, the alternative processes masks independently, which is less effective and also slower due to requiring multiple forward passes over a batch of masked inputs rather than a single image. Second, the pooling strategy can introduce information loss by collapsing spatial structure without explicitly modeling interactions among entities. Third, the number of entity embeddings varies across images, which makes optimization more challenging for self-attention, especially when positional encoding is required.

\begin{table}[t!]
\centering
\small
\caption{\textbf{Ablation study on alternative mask encoder/decoder designs for directly encoding binary entity masks in the panoptic narrative grounding task.} ``Conv'' indicate learnable convolution layer with kernel size $14\times 14$.}
\label{tab:mask_structure}
\begin{tabular}{ccc|c}
\cellcolor{lightgray!30} Mask Value Range & \cellcolor{lightgray!30} Pooling Style & \cellcolor{lightgray!30} Attention Blocks & \cellcolor{lightgray!30} AR \\ \toprule
$\{0,1\}$ & Avg & 1 & 60.6  \\
$\{-1,1\}$ & Avg & 1 & 60.7 \\
$\{0,255\}$ & Avg & 1 & 61.5 \\ 
$\{0,255\}$ & Conv & 1 & 61.2  \\
$\{0,255\}$ & CLS & 1 & 61.4  \\
$\{0,255\}$ & Avg & 2 & 61.9 \\
$\{0,255\}$ & Avg & 3 & 62.4 \\
$\{0,255\}$ & Avg & 4 & 62.4 \\
$\{0,255\}$ & Avg & 5 & 62.6 \\ \midrule
\multicolumn{3}{c}{ColorMap (ours)} & \textbf{69.8} \\ \bottomrule
\end{tabular}
\end{table}

Furthermore, we conduct another ablation on alternative mask designs to replace our proposed colormap encoder. These alternatives include directly extracting region features and incorporating mask weights into the attention scores of the cross-attention module between entity embeddings and image features. In Table~\ref{tab:mask_structure_2}, we show that our colormap design still outperforms the alternative designs, even when the best hyper-parameters of the other two designs are used. A possible reason is that the colormap representation enriches inter-entity cues in both the encoder and the decoder, whereas the other two alternatives introduce such interactions only in the decoder.

\begin{table}[t!]
\centering
\footnotesize
\caption{\textbf{Ablation Study on Mask Decoder Design in Panoptic Narrative Grounding Task.} For region feature extraction, we use the RoI-Align operation with bounding boxes on CLIP features to obtain the entity embeddings. For incorporating mask weights into the attention scores, we compare additive and multiplicative variants, where the entity embeddings are initialized from CropFormer and then refined through interleaved cross-attention with image features and self-attention among entity embeddings. `CA' indicates the cross-attention. }
\label{tab:mask_structure_2}
\begin{tabular}{c|c|c}
Mask Decoder Design & Operation  & AR \\ \toprule
Region Feature Extraction & RoI Align on Clip Feature & 62.9 \\ \midrule
\multirow{2}*{Incorporating mask weights into CA} & Adding Mask in CA & 63.2 \\ 
& Multiply Mask in CA & 60.3 \\ \midrule
ColorMap Encoder (ours) & - &\textbf{69.8} \\ \bottomrule
\end{tabular}
\end{table}

\vspace{1mm}
\noindent \textbf{Sampling Strategy.}
Table~\ref{tab:results_aba_png} shows how varying the sample ratio of three distinct subtasks—image description, semantic noun extraction, and entity grounding affects model performance. 
Notably, the entity grounding subtask includes a component that evaluates the effectiveness of noun extraction within the COCO panoptic narrative grounding dataset. 
Our initial approach, depicted in the first row, exclusively uses PNG data for training and performs well in entity grounding. 
Nevertheless, the Average Recall (AR) for semantic noun extraction under this setup is suboptimal, ultimately constraining the overall segmentation efficacy.
Incorporating captions generated by GPT-4 from the EntitySeg dataset enhances the language model accuracy in noun extraction. 
These results demonstrate that once the noun extraction subtask is included, the exact sample ratio becomes less critical, as no marked performance discrepancy is observed across different ratios. 
Additionally, integrating the image description task facilitates our framework to autonomously produce image captions.
With sample ratios of 0.2, 0.2, and 0.6 for image description, semantic noun extraction, and entity grounding, our GELLA framework obtains the best performance for both entity recognition and panoptic narrative grounding.

\begin{table}[t!]
\centering
\footnotesize
\vspace{-2mm}
\caption{\textbf{Ablation study of the sampling strategy in panoptic narrative grounding.} ``IMG DES'', ``NOUN EXT'', and ``ENT GRO'' indicate three sampled subtasks, including image description, semantic noun extraction, and entity grounding in our training.}
\vspace{-2mm}
\label{tab:results_aba_png}
\setlength\tabcolsep{2pt}
\begin{tabular}{ccccccccc}
\toprule
\cellcolor{lightgray!30}
& 
\cellcolor{lightgray!30}  
& 
\cellcolor{lightgray!30}  
& 
\multicolumn{5}{c}{\cellcolor{lightgray!30} Segmentation} & \multicolumn{1}{c}{\cellcolor{lightgray!30} Text}
\\
\multirow{-2}{*}{\cellcolor{lightgray!30} \makecell[c]{IMG\\DES}} & 
\multirow{-2}{*}{\cellcolor{lightgray!30} \makecell[c]{ NOUN\\EXT}} & 
\multirow{-2}{*}{\cellcolor{lightgray!30} \makecell[c]{ENT\\GRO}}
& 
\cellcolor{lightgray!30} AR & 
\cellcolor{lightgray!30} AR$_{\text{Th}}$ & 
\cellcolor{lightgray!30} AR$_{\text{St}}$ & \cellcolor{lightgray!30} AR$_{\text{Sing}}$ & \cellcolor{lightgray!30} AR$_{\text{Pl}}$ & 
\cellcolor{lightgray!30} AR$_\text{text}$ \\ \midrule
0.0 & 0.0 & 1.0 & 66.4 & 64.2 & 70.5 & 68.0 & 58.3 & 79.4 \\
0.0 & 0.2 & 0.8 & 69.4 & 65.8 & 71.8 & 69.5 & 58.0 & 95.4 \\
0.1 & 0.1 & 0.8 & 69.2 & 65.5 & 71.5 & 69.2 & 57.8 & 95.7 \\
0.2 & 0.2 & 0.6 & 69.8 & 66.2 & 72.3 & 69.9 & 58.4 & 96.1 \\ \bottomrule
\end{tabular}
\end{table}

\vspace{1mm}
\noindent \textbf{Comparison with Various Segmentation Inputs.} Table~\ref{tab:aba_seg_results} analyzes the effect of different class-agnostic segmentation inputs on panoptic narrative grounding. We compare four types of segmentation inputs: Mask2Former~\cite{cheng2022masked} trained on the class-aware panoptic segmentation on the COCO dataset~\cite{lin2014microsoft}, SAM~\cite{kirillov2023segment} trained on the class-agnostic SA-1B dataset~\cite{kirillov2023segment} with two hyperparameter settings, and CropFormer~\cite{qi2023high} trained on the EntitySeg dataset~\cite{qi2023high}. 
The results in Table~\ref{tab:aba_seg_results} demonstrate that EntitySeg segmentation inputs outperform the other three. 
This can be attributed to CropFormer's design for entity-level inputs, which align more closely with GELLA for grounding entities. 
In contrast, SAM, with its three-level predictions for each pixel, faces challenges in automatically generating entity-level segmentation results without user interaction, even with two suggested hyperparameter settings. 
These results underscore the significance of segmentation mask quality in strengthening GELLA's performance.

\begin{table}
\centering
\footnotesize
\setlength\tabcolsep{2pt}
\caption{\textbf{Ablation study of various colormap inputs on panoptic narrative grounding.} We use two hyperparameter settings for inference with SAM, denoted as ``SAM-1'' and ``SAM-2''. ``SAM-1'' adopts the default setting; the hyperparameters of ``SAM-2'' are on \href{https://github.com/facebookresearch/segment-anything/blob/main/notebooks/automatic_mask_generator_example.ipynb}{SAM's GitHub page}: 
\textit{points\_per\_side} (32), \textit{pred\_iou\_thresh} (0.86), \textit{stability\_score\_thresh} (0.92), \textit{crop\_n\_layers} (1), \textit{crop\_n\_points\_downscale\_factor} (2), \textit{min\_mask\_region\_area} (100).}
\label{tab:aba_seg_results}
\begin{tabular}{ccccccc}
\toprule
\cellcolor{lightgray!30}
& 
\cellcolor{lightgray!30}
&
\multicolumn{5}{c}{\cellcolor{lightgray!30} Segmentation}
\\
\multirow{-2}{*}{\cellcolor{lightgray!30} \makecell[c]{Structure}} & \multirow{-2}*{\cellcolor{lightgray!30} \makecell[c] { Backbone}} & \cellcolor{lightgray!30} AR & \cellcolor{lightgray!30} AR$_{\text{Th}}$ & 
\cellcolor{lightgray!30} AR$_{\text{St}}$ & \cellcolor{lightgray!30} AR$_{\text{Sing}}$ & \cellcolor{lightgray!30} AR$_{\text{Pl}}$ \\ \midrule
Mask2Former-PAN & Swin-L & 66.0 & 63.8 & 68.6 & 66.0 & 55.7\\
SAM-1 & VIT-H & 67.9 & 64.8 & 70.3 & 68.7 & 58.1 \\
SAM-2 & VIT-H & 66.1 & 63.9 & 68.7 & 67.0 & 57.6 \\
CropFormer & Swin-L & 69.8 & 66.2 & 72.3 & 69.9 & 58.4 \\ \bottomrule
\end{tabular}
\end{table}

\begin{table}[t!]
\centering
\footnotesize
\caption{\textbf{Ablation study on the association module design.}}
\label{tab:results_aba_asso}
\begin{tabular}{cccccc}
\toprule
\cellcolor{lightgray!30}
& 
\multicolumn{5}{c}{\cellcolor{lightgray!30} Segmentation}
\\
\multirow{-2}{*}{\cellcolor{lightgray!30} \makecell[c]{Structure}} & 
\cellcolor{lightgray!30} AR & 
\cellcolor{lightgray!30} AR$_{\text{Th}}$ & 
\cellcolor{lightgray!30} AR$_{\text{St}}$ & \cellcolor{lightgray!30} AR$_{\text{Sing}}$ & \cellcolor{lightgray!30} AR$_{\text{Pl}}$ \\ \midrule
FC+ReLU+FC & 69.7 & 66.1 & 72.3 & 69.7 & 58.2 \\
FC & 69.8 & 66.2 & 72.3 & 69.9 & 58.4 \\ \bottomrule
\end{tabular}
\end{table}

\noindent \textbf{Association Module.}
Table~\ref{tab:results_aba_asso} presents the ablation study focused on the structural design within our association module. 
The study reveals that incorporating additional fully connected (FC) layers does not yield further performance improvements. 
This plateau in enhancement can be attributed to the fact that the final embeddings of both the $<$SEG$>$ token and the entities already reside within the same domain of the CLIP image space. 
Consequently, there is no need for complex operations to align them within a shared space, as they are inherently congruent.

\subsection{Visualization}
Figure~\ref{fig.user_interface} displays the user interface of our GELLA framework, showcasing its capabilities in image captioning, entity recognition, and panoptic narrative grounding.
The GELLA model first describes the image in this interactive interface and outputs a long caption. 
Then, it extracts each semantic noun and associates it with the provided entity segmentation masks. 
Additionally, the GELLA framework can incorporate existing segmentation results as input. 
More results are available in the appendix.

\vspace{-4mm}
\subsection{Failure Case}
From the visualization results shown in the appendix, the final performance of our pipeline is influenced by each intermediate step, including entity segmentation, long caption generation, entity recognition, and association. 
Apart from the segmentation quality achieved by CropFormer, our GELLA model has three notable limitations or failure cases.
First, the extraction of semantic nouns may fail to include all ground truth nouns. 
For example, phrases like `in front of the \textit{sofa}' or `on the \textit{windowsill}' are missing in the first instance of Fig. 18 of our appendix.
Second, the generated captions do not consistently account for all the entity segmentation results. 
Lastly, our model currently cannot handle part segmentation or the subsequent association, which we plan to address in future work.

\section{Conclusion}
We propose ESG, a decoupled pipeline with two modules including CropFormer and GELLA for entity segmentation and grounding.
Specifically, the CropFormer framework, along with an fine-grained EntitySeg dataset, focus on in-the-wild generalization and high-quality dense segmentation. 
The EntitySeg dataset contains about 33K images from various domains and resolutions, featuring high-quality mask annotations for training and testing. 
The CropFormer framework is designed to address the computational challenges posed by the high-quality and high-resolution characteristics of the EntitySeg dataset.
Specifically, it enables fusing multi-view predictions from a low-res full image and its high-res crops by learning batch-level queries. 
Furthermore, we introduce GELLA, a framework that leverages a large language model to ground entities with long captions. 
The GELLA framework consists of colormap and image encoders, and mask and language decoders for association.
Compared to other works, the colormap encoder enables the network to prioritize masks. 
This allows us to use the CLIP vision encoder to handle low-resolution images in pixel-level prediction and semantic noun extraction with the language decoder. 
We hope that our proposed ESG can provide a more flexible pipeline to ground entities by receiving offline masks generated by state-of-the-art segmentation methods. 

\small{
        \vspace{-3mm}
	\bibliographystyle{IEEEtran}
	\bibliography{IEEEabrv,ref}
}

\begin{IEEEbiography}
[{\includegraphics[width=1in,height=1.25in,clip,keepaspectratio]
	{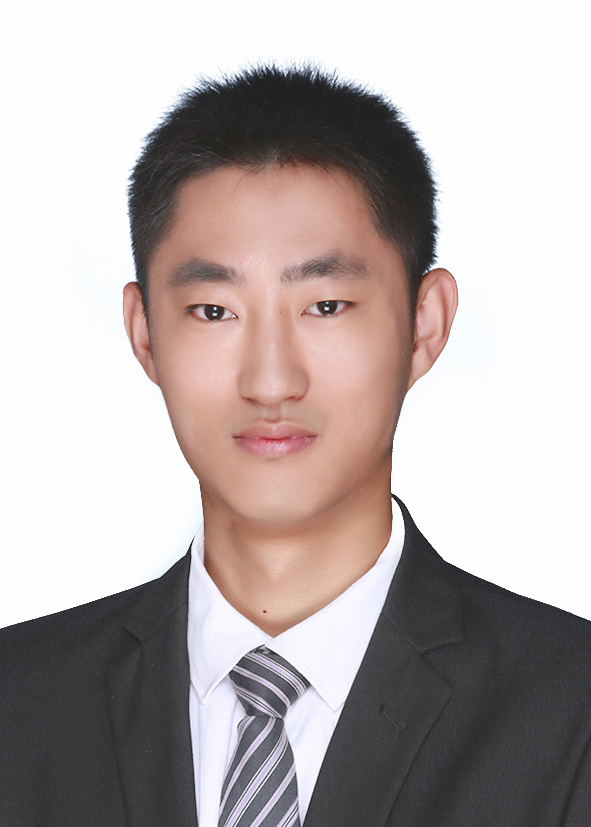}}]
        {Lu Qi} is a full professor of Wuhan University and the AI director of Insta360. Before that, he worked as a postdoctoral researcher at UC Merced and has over 17,000 citations in Google Scholar. He received the Ph.D. degree from The Chinese University of Hong Kong in 2021. His research interests include instance-level detection, image generation, and cross-modal pretraining. He is the associate editor of TPAMI. He was the senior program committee of AAAI 2023/2024/2025 and area chair of ICLR2024/2025, NeurlPS2024, ICML2025, and ICCV2025.
\end{IEEEbiography}

\begin{IEEEbiography}
	[{\includegraphics[width=1in,height=1.25in,clip,keepaspectratio]
		{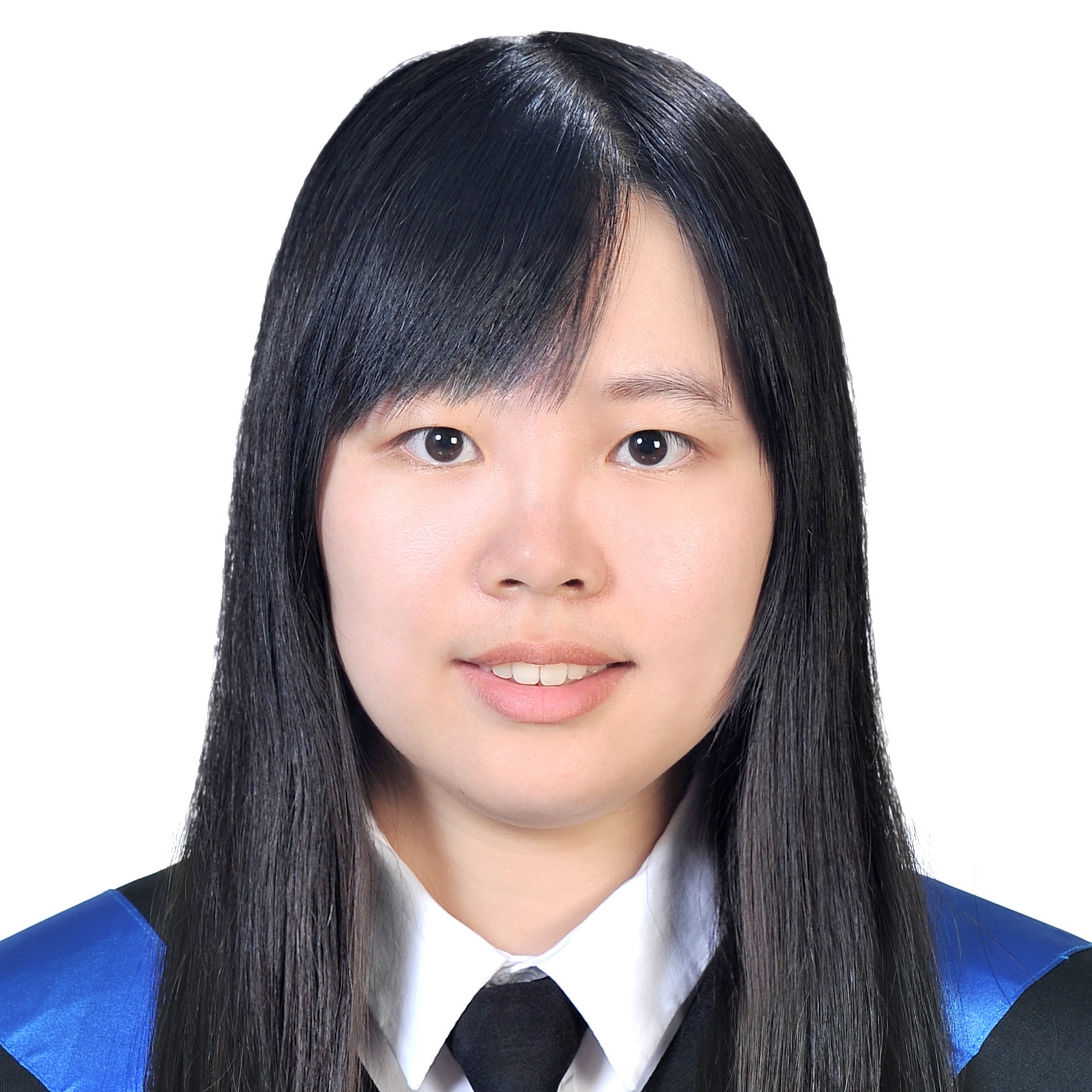}}]
	{Yi-Wen Chen}
	is a Ph.D. student in the Department of Electrical Engineering and Computer Science at the University of California, Merced, under the supervision of Prof. Ming-Hsuan Yang. Before that, she received her B.S. and M.S. degrees in Electrical Engineering from National Taiwan University in 2017 and 2019, respectively. Her research interests lie in computer vision and deep learning, primarily focusing on vision, language reasoning, and video understanding.
\end{IEEEbiography}

\begin{IEEEbiography}
[{\includegraphics[width=1in,height=1.25in,clip,keepaspectratio]{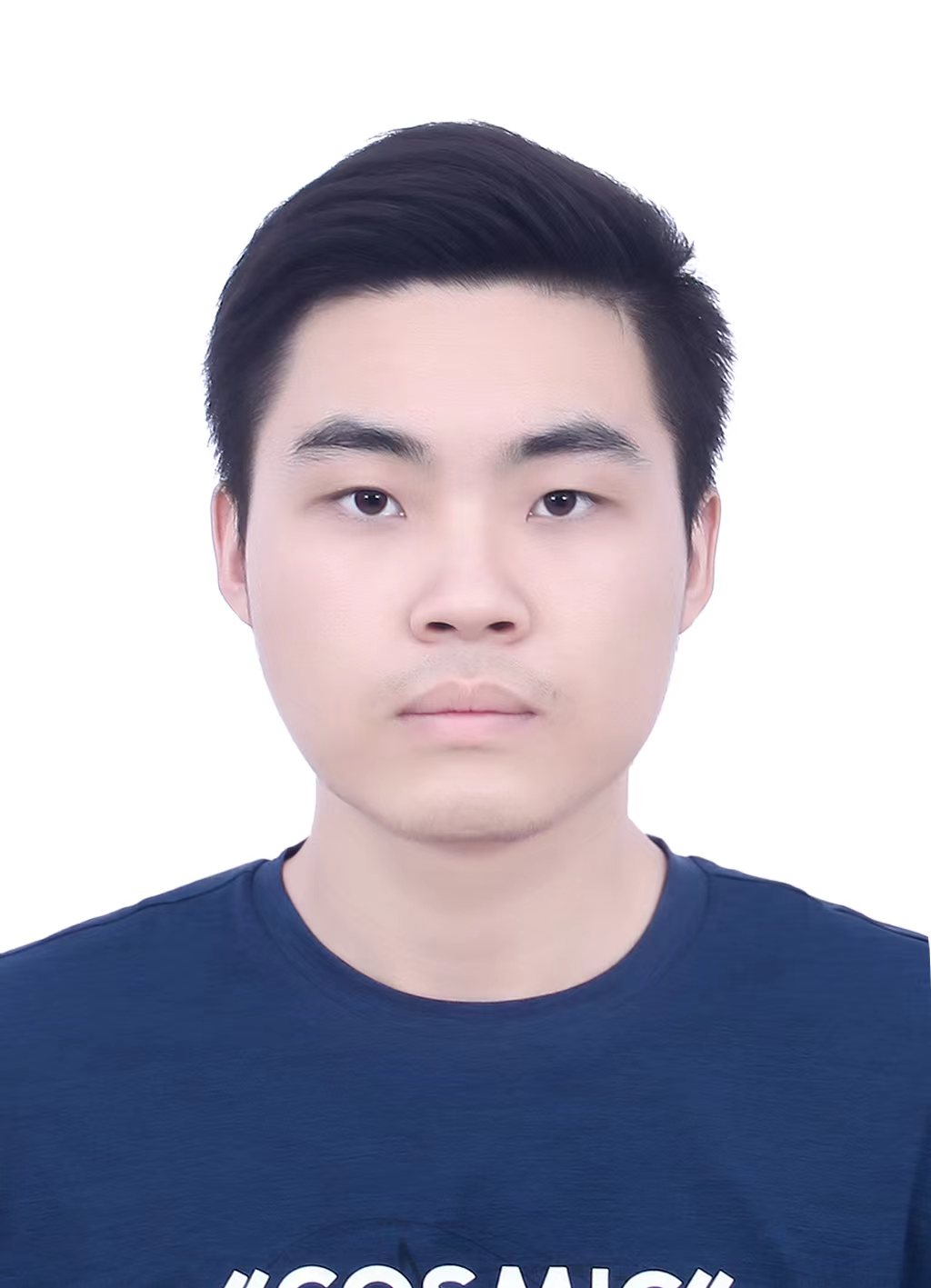}}]{Tao Zhang} received the Master degree from the School of Remote Sensing and Information Engineering at Wuhan University (WHU) in 2023. Currently, he is pursuing his PhD at Wuhan University. He served as reviewers for CVPR, ICCV, ECCV, NIPS, ICLR, AAAI, ACCV, TMM, TCSVT, etc. His research interests revolve around image understanding, video understanding, multi-modal large language model, and remote sensing.
\end{IEEEbiography}

\begin{IEEEbiography}[{\includegraphics[width=1in,height=1.25in,clip,keepaspectratio]{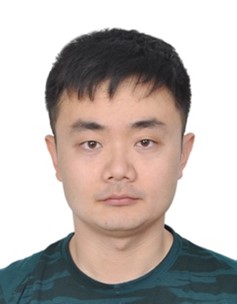}}]{Xiangtai Li} is working as a Research Fellow in S-Lab, a member of the Multimedia Laboratory of NTU (MMLab@NTU) at Nanyang Technological University. He received the Ph.D. degree from Peking University in 2022. His research interests include computer vision and machine learning with a focus on scene understanding, segmentation, video understanding, and multi-modal learning. Several of his works have been published in top-tier conferences and journals. He regularly reviews top-tier conferences and journals, including CVPR, ICCV, ICLR, ICML, NeurIPS, T-PAMI, and IJCV.
\end{IEEEbiography}

\begin{IEEEbiography}
[{\includegraphics[width=1in,height=1.25in,clip,keepaspectratio]
	{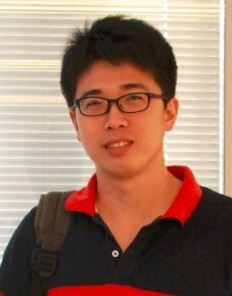}}]
{Xu Yang}
received a B.E. degree in electronic engineering from the Ocean University of China, Qingdao, China, in 2009, and a Ph.D. degree in computer science from the Institute of Automation, Chinese Academy of Sciences (CASIA), Beijing, China, in 2014. He is an associate Professor with the State Key Laboratory of Management and Control for Complex Systems, Institute of Automation, CASIA, Beijing. His current research interests include computer vision, graph algorithms, and robotics.
\end{IEEEbiography}
\vspace{-12mm}
\begin{IEEEbiography}
[{\includegraphics[width=1in,height=1.25in,clip,keepaspectratio]
	{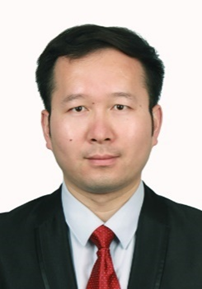}}]
{Bo Du} is currently a professor at the School of Computer Science and Institute of Artificial Intelligence of Wuhan University. He is also the director of the National Engineering Research Center for Multimedia Software, Wuhan University, Wuhan, China. He has more than 80 research papers published in the IEEE TPAMI, TIP, IEEE TCYB, IEEE TGRS, etc. Fourteen of them are ESI hot papers or highly cited papers. His major research interests include machine learning, computer vision, and image processing. He is currently a senior member of IEEE and serves as associate editor for Neural Networks, Pattern Recognition and Neurocomputing. He won the Highly Cited Researcher (2019/2020) by the Web of Science Group. He won IEEE Geoscience and Remote Sensing Society 2020 Transactions Prize Paper Award, the IJCAI (International Joint Conferences on Artificial Intelligence) Distinguished Paper Prize, IEEE Data Fusion Contest Champion, and IEEE Workshop on Hyperspectral Image and Signal Processing Best Paper Award.
\end{IEEEbiography}

\begin{IEEEbiography}
[{\includegraphics[width=1in,height=1.25in,clip,keepaspectratio]
	{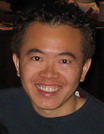}}]
{Ming-Hsuan Yang} is a professor of Electrical
Engineering and Computer Science at the University
of California, Merced. Yang serves as a program
co-chair of the IEEE International Conference on
Computer Vision (ICCV) in 2019, program cochair of the Asian Conference on Computer Vision(ACCV) in 2014, and general co-chair of ACCV 2016. Yang served as an associate editor of
the IEEE Transactions on Pattern Analysis and
Machine Intelligence from 2007 to 2011, and is
an associate editor of the International Journal of
Computer Vision, Image and Vision Computing, and Journal of Artificial Intelligence Research. He received the NSF CAREER award in 2012 and Google Faculty Award in 2009. He is a Fellow of the IEEE.
\end{IEEEbiography}

\end{document}

%% file: mlVecMat.tex
\usepackage{amsmath}
\usepackage{amsfonts}
\usepackage{amssymb}
\usepackage{wrapfig}
\usepackage{subcaption}
\usepackage{multirow}
 \usepackage{mathtools} 

\usepackage{verbatim}

\usepackage{anyfontsize}

\usepackage{microtype}
\usepackage{graphicx}
\usepackage{booktabs} 
\usepackage{multirow}
\usepackage{amsmath,amssymb}
\usepackage{booktabs}
\usepackage{caption,subcaption}

\usepackage{xcolor}
\definecolor{mygreen}{HTML}{3cb44b}
\definecolor{skyblue}{HTML}{beffff}
\definecolor{lightgreen}{HTML}{90ee90}

\usepackage{color, colortbl}

\definecolor{emerald}{rgb}{0.31, 0.78, 0.37}

\usepackage{tcolorbox}
\usepackage{enumitem}
\setitemize{itemsep=10pt,topsep=0pt,parsep=0pt,partopsep=0pt}
\usepackage{colortbl}

\usepackage{xcolor}
\definecolor{mygreen}{HTML}{3cb44b}
\colorlet{myyellow}{green!10!orange!90!}
\makeatletter

\usepackage{tikz}
\usetikzlibrary{arrows,shapes,snakes,automata,backgrounds,fit,petri}
\usepackage{adjustbox}

\newcommand{\RN}[1]{%
	\textup{\lowercase\expandafter{\it \romannumeral#1}}%
}
\usepackage{tabu}

\newcommand{\beq}{\vspace{0mm}\begin{equation}}
\newcommand{\eeq}{\vspace{0mm}\end{equation}}
\newcommand{\beqs}{\vspace{0mm}\begin{eqnarray}}
\newcommand{\eeqs}{\vspace{0mm}\end{eqnarray}}
\newcommand{\barr}{\begin{array}}
\newcommand{\earr}{\end{array}}

\usepackage{color, colortbl}
\definecolor{Gray}{gray}{0.93}

\usepackage{lipsum}

\usepackage{pifont}

\usepackage{makecell}

\usepackage{xcolor,amsmath}
\usepackage[linesnumbered,ruled,vlined]{algorithm2e}
\DontPrintSemicolon

\usepackage{xcolor}
\definecolor{mygreen}{HTML}{3cb44b}

\SetKwComment{Comment}{\color{green!50!black}\# }{}

\SetKwProg{Function}{def}{:}{}

\SetKwProg{For}{for}{:}{}
\SetKwProg{If}{if}{:}{}
\newcommand{\VarSty}[1]{\textnormal{\ttfamily\color{blue!90!black}#1}\unskip}